\documentclass[preprint,11pt]{elsarticle}

\usepackage[top=1in, bottom=1in, left=1in, right=1in]{geometry}

\usepackage{array}
\usepackage{enumitem}
\usepackage{esvect}
\usepackage{natbib}
\usepackage[toc,page]{appendix}
\usepackage{comment}
\usepackage{multirow}

\usepackage[caption=false]{subfig}
\usepackage{pgfplots}

\usepackage[colorlinks=true, bookmarksopen,
            pdfsubject={algorithms},
            linkcolor={blue},
            anchorcolor={black},
            citecolor={black},
            filecolor={magenta},
            menucolor={black},
            plainpages=false,pdfpagelabels,
            urlcolor={blue}]{hyperref}

\usepackage{algorithmic}

\usepackage{graphicx,epstopdf}

\usepackage{enumitem}
\usepackage{subfig}
\usepackage{amsmath}
\usepackage{amssymb}
\usepackage{booktabs}
\usepackage{graphicx}
\usepackage{dcolumn}
\usepackage{mathtools}
\usepackage{amsmath,amsfonts}
\usepackage{bm}
\usepackage{amsmath}
\usepackage{amssymb}
\usepackage{color}
\usepackage{float}
\usepackage{setspace}
\usepackage{tabularx}
\usepackage{subcaption}

\allowdisplaybreaks

\newcommand{\be}{\begin{equation}}
\newcommand{\ee}{\end{equation}}

\makeatletter
\DeclareRobustCommand{\cev}[1]{%
  \mathpalette\do@cev{#1}%
}
\newcommand{\do@cev}[2]{%
  \fix@cev{#1}{+}%
  \reflectbox{$\m@th#1\vec{\reflectbox{$\fix@cev{#1}{-}\m@th#1#2\fix@cev{#1}{+}$}}$}%
  \fix@cev{#1}{-}%
}
\newcommand{\fix@cev}[2]{%
  \ifx#1\displaystyle
    \mkern#23mu
  \else
    \ifx#1\textstyle
      \mkern#23mu
    \else
      \ifx#1\scriptstyle
        \mkern#22mu
      \else
        \mkern#22mu
      \fi
    \fi
  \fi
}

\makeatother

\def\ba{\begin{array}}                \def\ea{\end{array}}
\def\bel{\begin{equation}\label}      \def\ee{\end{equation}}

\colorlet{texcscolor}{blue!50!black}
\colorlet{texemcolor}{red!70!black}
\colorlet{texpreamble}{red!70!black}
\colorlet{codebackground}{black!25!white!25}

\begin{document}

\title{Diffusion-based supervised learning of generative models for efficient sampling of multimodal distributions}

\author[inst1]{Hoang Tran}
\author[inst1]{Zezhong Zhang}
\author[inst2]{Feng Bao}
\author[inst3]{Dan Lu}
\author[inst1]{Guannan Zhang\corref{cor}}

\affiliation[inst1]{organization={Computer Science and Mathematics Division},
            addressline={Oak Ridge National Laboratory}, 
            city={Oak Ridge},
            state={TN},
            postcode={37831},
            country={USA}}

\affiliation[inst2]{organization={Department of Mathematics},
            addressline={Florida State University}, 
            city={Tallahassee},
            state={FL},
            postcode={32306}, 
            country={USA}}

\affiliation[inst3]{organization={Computational Science and Engineering Division},
            addressline={Oak Ridge National Laboratory}, 
            city={Oak Ridge},
            state={TN},
            postcode={37831}, 
            country={USA}}

\begin{frontmatter}
\begin{abstract}
We propose a hybrid generative model for efficient sampling of high-dimensional, multimodal probability distributions for Bayesian inference. Traditional Monte Carlo methods, such as the Metropolis-Hastings and Langevin Monte Carlo sampling methods, are effective for sampling from single-mode distributions in high-dimensional spaces. However, these methods struggle to produce samples with the correct proportions for each mode in multimodal distributions, especially for distributions with well separated modes. To address the challenges posed by multimodality, we adopt a divide-and-conquer strategy. We start by minimizing the energy function with initial guesses uniformly distributed within the prior domain to identify all the modes of the energy function. Then, we train a classifier to segment the domain corresponding to each mode. After the domain decomposition, we train a diffusion-model-assisted generative model for each identified mode within its support. Once each mode is characterized, we employ bridge sampling to estimate the normalizing constant, allowing us to directly adjust the ratios between the modes. Our numerical examples demonstrate that the proposed framework can effectively handle multimodal distributions with varying mode shapes in up to 100 dimensions. An application to Bayesian inverse problem for partial differential equations is also provided. 
\end{abstract}


\cortext[cor]{Corresponding author}
\tnotetext[fn1] {This manuscript has been authored by UT-Battelle, LLC, under contract DE-AC05-00OR22725 with the US Department of Energy (DOE). The US government retains and the publisher, by accepting the article for publication, acknowledges that the US government retains a nonexclusive, paid-up, irrevocable, worldwide license to publish or reproduce the published form of this manuscript, or allow others to do so, for US government purposes. DOE will provide public access to these results of federally sponsored research in accordance with the DOE Public Access Plan.}

\end{frontmatter}

%





%
%
%
%



\section{Introduction}
\label{sec:intro}

Sampling from an unnormalized, high-dimensional density function presents significant challenges in statistics and computational science. This problem is exacerbated in the case of multi-modal distributions, where probability mass is concentrated in multiple, potentially distant regions of the sample space. Multimodal distributions arise in situations where data naturally clusters around multiple centers or results from a combination of different statistical processes. Examples include mixture models like Gaussian mixture models, which combine several normal probability distributions to form a multimodal distribution; distributions with periodic behavior stemming from cyclic physical phenomena; and empirical data with multiple clusters due to underlying subpopulations or categories. Sampling from such distributions presents unique challenges, particularly due to isolated modes that are difficult to traverse between, local traps that can bias sampling algorithms, and increased complexity in high-dimensional spaces.

Several methods have been developed for sampling from complex distributions, including Markov Chain Monte Carlo (MCMC), nested sampling, and machine learning-based techniques. MCMC algorithms, such as random-walk Metropolis \cite{10.1063/1.1699114} and Gibbs sampling \cite{4767596}, generate samples by constructing a Markov chain whose stationary distribution is the target distribution. These methods have the tendency to explore parameter space via inefficient random walks, thus may take an unbearably long time to converge or fail to converge altogether, especially in multimodal scenarios where the sampling process can become trapped in certain modes. To remedy this issue, gradient-based sampling methods, such as Stochastic Gradient Langevin Dynamics \cite{welling2011bayesian} and Hamiltonian Monte Carlo \cite{neal2011mcmc,10.5555/2627435.2638586}, leverage gradient information for faster convergence and improved exploration. However, they still struggle to transition between well-separated modes in multimodal distributions as gradients tend to guide samples toward local modes. These methods are also highly sensitive to hyperparameters like step size and trajectory length, requiring careful tuning for optimal performance. 
Replica Exchange Monte Carlo \cite{swendsen1986replica,earl2005parallel}, also known as parallel tempering, improves multimodal exploration by running multiple chains at different temperatures and allowing swaps between them. While effective at overcoming energy barriers, this approach suffers from high computational costs due to the need for multiple parallel chains and careful temperature selection. Another popular approach for sampling multimodal posteriors with complex topologies is nested sampling \cite{feroz2009multinest, 10.1093/mnras/staa278, Buchner2021}, which, unlike traditional MCMC methods, adaptively refines its sampling strategy using a set of ``live points'' that improves efficiency by allocating more samples to high-likelihood regions. However, its computational cost scales poorly with high-dimensional problems, and the sampling process can be very slow.


A fundamental limitation shared by all above traditional probabilistic sampling methods is that they require restarting the algorithm each time new samples are needed. Recent advances in machine learning have spurred the development of data-driven sampling techniques that train deep neural samplers to approximate target distributions. Notable examples include normalizing flows \cite{kobyzev2020normalizing, wu2020stochastic}, diffusion models \cite{10.1145/3626235}, generative adversarial networks \cite{8253599}, and energy-based models \cite{lecun2006tutorial, du2019implicit}. Once trained, these models enable rapid sampling without the need to reinitialize the process, making them attractive alternatives to traditional methods. However, these approaches come with several limitations. They typically demand large amounts of high-quality training data and require careful architectural design and hyperparameter tuning to ensure stable training and avoid issues such as mode collapse. Furthermore, while they have demonstrated impressive empirical performance, they often lack rigorous theoretical guarantees of convergence. Additional challenges include high computational costs during training, sensitivity to initialization, and potential difficulties in generalizing to out-of-distribution data. Addressing these challenges remains an active area of research in the field.

In this work, we introduce a novel generative framework that overcomes several limitations of both traditional probabilistic methods and recent data-driven approaches for sampling from high-dimensional, multimodal distributions. Instead of tackling the entire distribution at once, our approach partitions the sampling task into smaller, more manageable problems. First, it detects distinct peaks in the target density and divides the input space into regions that each contain a single mode. Within each region, we use stochastic dynamics to generate preliminary samples and employ diffusion-based techniques to create labeled training pairs that map standard Gaussian inputs to the target distribution samples. With the availability of training data and the task now reduced to unimodal sampling, we employ a supervised training procedure, where the neural network models can be as simple as fully connected networks, whose architecture does not require excessive tuning and is robust to the problem. Finally, by estimating normalizing constants for each subdomain, we seamlessly integrate the individual samplers for each unimodal component into a unified generator for the target distribution. This strategy enables near-instantaneous sample generation while also overcoming limitations of modern deep learning approaches, such as the need for extensive training data and complex architecture design.

The paper is organized as follows. Section \ref{sec:prob} discusses the problem setting. In Section \ref{sec:method}, we present our methodology in detail, and Section \ref{sec:num} showcases numerical experiments that demonstrate the advantages of our approach compared to other baselines.

\section{Problem setting}\label{sec:prob}
This effort attacks the fundamental challenge of generating samples from high-dimensional multimodal probability distributions where only an unnormalized density function is available. 
In mathematical terms, the goal is to generate unrestricted number of samples of a target $d$-dimensional random variable $X \in \mathbb{R}^d$, characterized by an unknown probability density function (PDF). The density function $\pi(x)$ is defined by
\begin{equation}\label{eq:target}
     \pi(x) := \frac{\rho(x)}{\Lambda}\; \text{ with }\;
     \Lambda = \int_{\mathbb{R}^d} \rho(x) dx,
\end{equation}
where $\rho(x)$ is the unnormalized PDF, and $\Lambda$ represents the normalizing constant, an integral over $\mathbb{R}^d$
that is generally intractable due to high dimensionality. We are particularly interested in sampling from multimodal distributions defined by the following general formulation:
\begin{equation}\label{eq:mm}
    \rho(x) := \sum_{k=1}^K r_k\, \rho_k (x),
\end{equation}
where $\rho_k(x)$ and $r_k$ denote the $k$-th mode of the target distribution and its weight, respectively. In this work, we do not assume $\{\rho_k (x)\}_{k=1}^K$ are identical modes, which makes the sampling problem more difficult. 

The objective of this work is to construct a parameterized generative model, denoted by
\begin{equation}\label{eq:transport}
    X = F(Y; \theta)\; \text{ with }\; Y \in \mathbb{R}^d,
\end{equation}
which maps a reference variable $Y$ following the standard Gaussian distribution to the target random variable $X$. This model enables sampling by mapping samples from $Y$ through the trained function 
$F$, with the parameter $\theta$ optimized to approximate the distribution $\pi(x)$ as closely as possible. The model's output should accurately reproduce samples of 
$X$ that reflect the true distribution, even though traditional sample-based training is infeasible. We emphasize that the problem setup is different from traditional generative modeling because no training data or samples of 
$X$ are available for model training. Instead, only the function $\rho(x)$, which can be evaluated at any point in $\mathbb{R}^d$, is available. Thus, the challenge becomes developing a generative model that can accurately approximate 
$\pi(x)$ based solely on queries to the unnormalized multimodal probability distribution $\rho(x)$.



\section{Diffusion-based generative models for sampling multimodal distributions}\label{sec:method}

This section provides a detailed description of the proposed methodology. Sampling from complex, high-dimensional, and multimodal distributions is a fundamental challenge in probabilistic modeling and Bayesian inference. We introduce a novel approach that leverages a divide-and-conquer strategy to address this challenge. Our approach decomposes the problem into manageable subproblems by first identifying all potential modes of the target distribution, thereby transforming the global sampling task into a set of simpler, unimodal sampling problems. We then develop a supervised learning-based methodology to generate samples efficiently: for each unimodal component, we construct and train a model that maps standard Gaussian samples to the target distribution. Finally, we estimate the appropriate mixing ratios by computing the normalizing constants of each unimodal mode, ensuring an accurate representation of the overall distribution. Our framework is modular, allowing each step to be adapted independently based on specific application needs. In the following sections, we provide a detailed exposition of each component of our framework.
Specifically, the steps are as follows:
%
\begin{itemize}[leftmargin=20pt]
\item {\bf Step 1:} Identify the number of modes and segment the input domain into multiple subdomains, each of which contains a single mode of the target distribution (Section \ref{sec:opt}).
\item {\bf Step 2:} Generate samples for each unimodal component of the target distribution within the corresponding subdomain using Lagenvin dynamics (Section \ref{sec:lang}).
\item {\bf Step 3:} Generate labeled data between samples from the standard Gaussian distribution and each mode of the target distribution using training-free diffusion models (Section \ref{sec:diff}).
\item {\bf Step 4:} Train
a fully-connected neural network for each unimodal component of the target distribution using the generated labeled data (Section \ref{sec:NN}).
\item {\bf Step 5:} Calculate the mixing ratios of the unimodal components of the target distribution by estimating their normalizing constants (Section \ref{sec:bridge_sampling}).
\item {\bf Step 6:} Assemble the final generator from the trained models and generate new samples (Section \ref{sec:gen_sample}).  
\end{itemize}
The resulting generative model enables near-instantaneous sampling from complex, multimodal distributions, providing an effective and scalable solution to a longstanding problem in high-dimensional sampling. The details of each step are discussed in the following subsections.

\subsection{Mode identification and domain segmentation for unimodal decomposition}
\label{sec:opt}

We employ a multi-start optimization procedure to identify the modes of the unnormalized probability density function (PDF) $\rho(x)$ by locating its peaks. The objective is to explore the local maxima of $\log \rho(x)$ when $\rho(x)$ exhibits multiple modes. However, due to the nonconvexity of $\log \rho(x)$, gradient-based methods often converge to a single local maximum, making the outcome highly sensitive to the choice of the initial point $x_0$.
In the multi-start approach, the optimization process begins with $N$ initial points, $\{x_{0,n}\}_{n=1}^N$, sampled from a prior distribution $\mathcal{P}(x)$. For each starting point $x_{0,n}$, a local optimization method, such as gradient descent, is applied to obtain a local maximum $x_n^{*}$: 
\begin{align*}
    x_{n}^{*} = \mathcal{M}\left(x_{0,n}\right),
\end{align*}
where $\mathcal{M}$ denotes the operation of a first order or quasi-Newton optimization method such as the BFGS algorithm. Given the energy function $E(x) = -\log {\rho}(x)$, we use the standard gradient descent approach with fixed step size $\lambda$: 
\begin{align}
\label{eq:GD}
x_{j+1,n} = x_{j,n} - \lambda \nabla E(x_{j,n}).
\end{align}
The output of interest at this step is a collection of local maxima $\{x^*_1,x^*_2,\ldots,x^*_K\}$, each of which represents the peak of a possible mode of the target distribution $\rho(x)$. The success of the multi-start method depends on several factors, such as the number of initial points $N$ and the distribution $\mathcal{P}(x)$, which controls the exploration of the input space, as well as the quality and convergence properties of the local optimizer $\mathcal{M}$. 

Next, we divide the set of initial points $\{x_{0,n}\}_{n=1}^N$ into $K$ groups based on the corresponding optima reached by Eq.~\eqref{eq:GD} when initialized at each point. We segment the domain $\mathbb{R}^d$ into $K$ distinct subregions $\Omega_1,\ldots,\Omega_K$ according to these groups using C-Support Vector Classification (C-SVC), i.e.,
\[
\Omega_{1}\cup \ldots \cup \Omega_K = \mathbb{R}^d,\ \ \Omega_{i}\cap \Omega_{j} = \emptyset,\, \forall i\ne j. 
\]
This method constructs decision boundaries by solving an optimization problem that maximizes the margin between groups while allowing for some misclassification through the introduction of slack variables, thereby extending the binary Support Vector Machine framework using strategies such as the one-vs-one or one-vs-rest approaches. To handle nonlinear separations, kernel functions are used to implicitly map the input data into a higher-dimensional feature space where the decision boundaries become linear. Common kernels include the radial basis function (RBF), sigmoid, and polynomial kernels. For more details about the C-SVC method, we refer to \cite{platt1999probabilistic}. We define $\hat{\rho}_k$ as the restriction of $\rho$ within each subdomain $\Omega_k$.
\begin{equation}\label{eq:rhok}
    \hat{\rho}_k = \rho|_{\Omega_k}, \ k = 1,\ldots,K.
\end{equation}
By construction, $\hat{\rho}_k$ are unimodal or nearly unimodal functions. We note that it is not necessary that $\hat{\rho}_k = {\rho}_k$, where $\rho_k$ is the $k$-th mode of the target distribution (see definition in \eqref{eq:mm}). Indeed, this relation holds only when the supports of $\rho_k$ do not overlap, and we do not rely on it in our development. In the next steps (Sections \ref{sec:lang}--\ref{sec:NN}), we break down the problem of sampling from $\rho$ into a set of easier problems of sampling from $\hat{\rho}_k$ to construct our generative model. 

\subsection{Sampling unimodal components via Langevin dynamics}
\label{sec:lang}
We apply the Langevin dynamics to generate samples from unnormalized, nearly unimodal distributions $\hat{\rho}_1, \hat{\rho}_2, \ldots, \hat{\rho}_K$, obtained using the algorithm in Section \ref{sec:opt}. Langevin dynamics combines gradient-based deterministic updates with stochastic noise to generate samples for each $\hat{\rho}_k(x)$. In particular, the updates are governed by the following stochastic differential equation (SDE):
\[
\mathrm{d}x_t = -\nabla E_k(x_t) \, \mathrm{d}t + \sqrt{2} \, \mathrm{d}W_t,
\]
where $E_k$ is the energy function $E_k(x ) = -\log \hat{\rho}_k(x)$, $\nabla E_k$ is the score function, $t$ denotes time, and $W_t$ is a Wiener process modeling the Gaussian noise. 

We discretize the SDE using the Euler-Maruyama scheme with a step size $\eta$:
\begin{align}
\label{eq:lgv}
x_{j+1} = x_j - \eta \nabla E_k(x_j) + \sqrt{2\eta} \, \xi_j,
\end{align}
where $\xi_j \sim \mathcal{N}(0, I)$ is sampled from a standard multivariate Gaussian distribution. The first term, $- \eta \nabla E_k(x_j)$, drives the system towards regions of higher probability, while the noise term $\sqrt{2\eta} \, \xi_j$ ensures exploration of the state space. Under suitable conditions, the sequence $\{x_j\}$ generated by this process converges to the target distribution $\hat{\rho}_k(x)$ as $j \to \infty$. Proper tuning of the step size $\eta$ and sufficient iterations are critical to achieving accurate approximations of $\hat{\rho}_k(x)$. The classifier based on C-SVC is used to constrain the Langevin dynamics within each subdomain $\Omega_k$. The outcome of this step is a set of samples of each mode $\hat{\rho}_k(x)$ of the target distribution, denoted by
\begin{equation}\label{eq:unisamples}
    \{x_{k,m}, m = 1,\ldots,M\} \;\;\text{ for }\; k = 1, \ldots, K,
\end{equation}
where $M$ is the number of samples from each mode and $K$ is the number of identified modes of the target distribution. 

We emphasize that sampling from a unimodal distribution using Langevin dynamics is significantly simpler than sampling from a multimodal distribution due to the nature of the energy landscape and the behavior of the stochastic dynamics. When dealing with a multimodal distribution, multiple peaks exist, separated by regions of low probability. Langevin dynamics often struggles to transition between these modes due to energy barriers, causing the sampler to become trapped in a single local mode for extended periods. Overcoming these challenges in the multimodal case typically requires additional techniques such as tempered transitions, replica exchange methods, whereas such modifications are unnecessary for unimodal distributions. Moreover, Langevin dynamics struggles to estimate the correct mixing ratio of multimodal distributions because it tends to get trapped in individual modes, leading to biased sampling that does not accurately reflect the relative probabilities of different modes. In our method, we do not rely on Langevin dynamics to estimate the mixing ratio of different modes, so we could have the same number of samples generated for each unimodal component. The mixing ratio is estimated in Section \ref{sec:bridge_sampling}.

\subsection{Generating labeled data via training-free diffusion models}
\label{sec:diff}
Labeled data is critical to train a generative model in supervised manner. 
Here, we employ a score-based diffusion model \citep{song2021scorebased,pmlr-v115-song20a} to transform samples from a simple standard distribution into those following the unimodal distribution $\hat{\rho}_k$. Similar to Langevin sampling in Section \ref{sec:lang}, we build one diffusion model for each unimodal component of the target distribution. 
A score-based diffusion model consists of a forward stochastic differential equation (SDE) and a corresponding reverse probability flow ordinary differential equation (ODE), both defined within the standard pseudo-temporal domain $[0,1]$. The forward SDE, given by
\begin{equation}\label{eq:forward}
d Z_t = b(t) Z_{t} dt + \sigma(t) dW_t\; \text{ with }\; Z_0 = X \text{ and } Z_1 = Y,
\end{equation}
is used to map the target random variable $X \sim \hat{\rho}_k$, following the unimodal component $\hat{\rho}_k(x)$ in Eq.~\eqref{eq:rhok}, to the standard Gaussian random variable $Y \sim \mathcal{N}(0, \mathbf{I}_d)$ (i.e., the terminal state $Z_1$). There is a number of choices for the drift and diffusion coefficients in Eq.~\eqref{eq:forward} to ensure that the terminal state $Z_1$ follows $\mathcal{N}(0, \mathbf{I}_d)$ (see  \citep{song2021scorebased,10.1162/NECO_a_00142,NEURIPS2020_4c5bcfec,lu2022dpmsolver} for details). In this work, we choose $b(t)$ and $\sigma(t)$  in Eq.~\eqref{eq:forward} as 
\begin{equation}\label{eq:cof}
\begin{aligned}
b(t) = \frac{{\rm d} \log \alpha_t}{{\rm d} t} \;\;\; \text{ and }\;\;\; \sigma^2(t) = \frac{{\rm d} \beta_t^2}{{\rm d}t} - 2 \frac{{\rm d}\log \alpha_t}{{\rm d}t} \beta_t^2,
\end{aligned}
\end{equation}
where the two processes $\alpha_t$ and $\beta_t$ are defined by
\begin{equation}\label{eq:ab}
\alpha_t = 1-t, \;\; \beta_t^2 = t \;\; \text{ for } \;\; t \in [0,1].
\end{equation}
Because the forward SDE in \eqref{eq:forward}
is linear and driven by an additive noise, the conditional probability density function $q(Z_t | Z_0)$ for any fixed $Z_0$ is a Gaussian. In fact, 
\begin{equation}\label{eq:gauss}
q(Z_t | Z_0) = \mathcal{N}(\alpha_t Z_0, \beta_t^2 \mathbf{I}_d), 
\end{equation}
which leads to $q(Z_1 | Z_0) = \mathcal{N}(0, \mathbf{I}_d)$. The forward SDE defines a probability flow from $t=0$ to $t=1$, i.e., the evolution of the probability density function $q(t, Z_t)$ of the state $Z_t$ in Eq.~\eqref{eq:forward}. To generate new samples of $Z_0$, we need a reverse probability flow from $t=1$ to $t=0$. This can be achieved by solving the reverse ODE defined by
\begin{equation}\label{DM:RSDE}
d{Z}_t = \left[ b(t){Z}_t - \frac{1}{2}\sigma^2(t) S(Z_t, t)\right] dt \; \text{ with }\; Z_0 = X \text{ and } Z_1 = Y,
\end{equation}
where $S(Z_t, t)$ is the {score function}:
\begin{equation}\label{eq:exact_score}
 S(Z_{t}, t) = \nabla_z \log q(t,{Z}_t),
\end{equation}
where $q(t,Z_t)$ is the probability density function of $Z_t$ in Eq.~\eqref{eq:forward}. Compared to the reverse SDE, the reverse ODE defines a smooth function relationship between the initial state $Z_0$ and the terminal state $Z_1$, which can be used to generated labeled data to train a generator for each unimodal component in a supervised learning manner \cite{JMLMC_2024_Diffusion}. 

In this work, instead of using neural network to learn the score function $S(t,Z_t)$ in Eq.~\eqref{eq:exact_score}, we use the following Monte Carlo estimator of the score function
\begin{equation}\label{eq:MC}
S(Z_t, t) \approx \bar{S}(Z_t, t) :=  \sum_{m=1}^{M} - \frac{Z_t - \alpha_t x_{k,m}}{\beta^2_t} \bar{w}_t({Z_t},  x_{k,m}), 
\end{equation}
where $\{x_{k,m}\}_{m=1}^M$ is the set of samples generated by the Langevin dynamics for the 
unimodal component $\hat{\rho}_k(x)$ of the target distribution ${\rho}(x)$, and the weight function $w_t({Z_t},x_{k,m})$ is calculated by
\begin{equation}\label{eq:weight_app}
w({Z_t}, x_{k,m}) \approx  \bar{w}_t({Z_t}, x_{k,m}) := \frac{q(Z_t |  x_{k,m}) }{\sum_{m'=1}^{M} q(Z_t|  x_{k,m'})},
\end{equation}
and $q(Z_t |  x_{k,m})$ is the Gaussian distribution given in Eq.~\eqref{eq:gauss}. We refer to \cite{JMLMC_2024_Diffusion} for the derivation of the Monte Carlo estimator. In this way, we can solve the reverse ODE in Eq.~\eqref{DM:RSDE} directly using the dataset generated by the Langevin dynamics. The outcome of this step is a set of labeled data denoted by
\begin{equation}\label{eq:label_data}
    \mathcal{D}^{\rm label}_k := \left\{y_{k,j}, x_{k,j}\right\}_{j=1}^J\;\; \text{ for }\; k = 1, \ldots, K,
\end{equation}
for the $k$-th unimodal component, where 
$\{y_{k,j}\}_{j=1}^J$ are samples from the standard Gaussian distribution, $x_{k,j}$ is generated by solving the reverse ODE with the terminal condition being $y_{k,j}$. Note that the number of labeled data $J$ could be larger than the number of samples generated by the Langevin dynamics.

\subsection{Supervised training of neural networks for the unimodal components}
\label{sec:NN}
We conduct supervised training of a fully connected  neural network to learn a generator of each unimodal component $\hat{\rho}_k(x)$ of the target distribution $\rho(x)$ using the labeled data in Eq.~\eqref{eq:label_data}. The network defines a mapping function $T_k: \mathbb{R}^d \to \mathbb{R}^d$, parameterized by weights $\theta_k$, which maps $y_{k,j}$ into a predicted output $\hat{x}_{k,j} = T_k(y_{k,j}, \theta_k)$.
We train $T_k$ to minimize the mean squared error (MSE) loss measuring the discrepancy between $\hat{x}_{k,j}$ and the true target points $x_{k,j}$:
\begin{equation}\label{eq:mse}
    \mathcal{L}(\theta) = \frac{1}{J} \sum_{j=1}^J \|x_{k,j} - \hat{x}_{k,j}\|^2_2 = \frac{1}{J} \sum_{j=1}^J \|x_{k,j} - T_k(y_{k,j},\theta_k)\|^2_2. 
\end{equation}
The training is conducted by iteratively updating the network parameters $\theta$ using gradient-based optimization (e.g., stochastic gradient descent or Adam). This approach enables a supervised learning of the generative model, leveraging labeled data and overcoming multiple computational issues of unsupervised approaches. Once trained, the model can easily generate a large number of samples from the target distributions $\hat{\rho}_k$ at negligible cost, using randomly sampled inputs from a standard normal distribution.

\subsection{Estimating mixing ratios using Gaussian bridge sampling}
\label{sec:bridge_sampling}
In the last step, we employ the
bridge sampling method \cite{BENNETT1976245, meng1996simulating, Wang2016WarpBS} to quantify the contribution of each unimodal component $\hat{\rho}_k$ to the multimodal target distribution $\rho(x)$ and correctly allocate samples to each component $\hat{\rho}_1,\ldots, \hat{\rho}_K$. From the definition of $\hat{\rho}_k$, we can write $\rho(x) := \sum_{k=1}^K \hat{r}_k\, \hat{\rho}_k (x)$, where $\hat{r}_k$ are unknowns to be found. Note that this representation of $\rho$ is generally different from \eqref{eq:mm}, except when the supports of $\rho_k$ are non-overlapping. Bridge sampling leverages a proposal distribution to connect the target distribution and an auxiliary distribution that approximates $\rho(x)$ locally. 
We intend to estimate the normalizing constants $\Lambda_k$ of the unnormalized, local probability density functions $
\hat{\rho}_k$, expressed as
\[
\Lambda_k = \int_{\mathbb{R}^d} {\hat{\rho}}_k(x) \, \mathrm{d}x, 
\]
then the mixing ratio $\hat{r}_k$ for $\hat{\rho}_k$ can be computed by
\begin{align}
\hat{r}_k = \frac{\Lambda_k}{\sum_{k=1}^K \Lambda_k} .
\label{eq:rk}
\end{align}
To estimate $\Lambda_k$, Gaussian bridge sampling uses a Gaussian distribution $\phi$ with known normalizing constant $\Lambda_k$ as the proposal distribution and the simple identity 
\begin{align}
\label{eq:bridge_1}
\frac{\Lambda_k}{\Lambda_\phi} = \frac{\mathbb{E}_\phi[\hat{\rho}_k(x)\alpha(x)]}{\mathbb{E}_{\hat{\rho}_k}[\phi(x)\alpha(x)]}, 
\end{align}
which holds for any function $\alpha $, assuming that the common support of $\hat{\rho}_k$ and $\phi$ is non-trivial. Here, $\alpha$ serves as a ``bridge'' connecting $\hat{\rho}_k$ and $\phi$. The Monte Carlo method is then used to approximate $\Lambda_k/\Lambda_\phi$ in Eq.~\eqref{eq:bridge_1}, i.e., 
\begin{align}
\label{eq:bridge_2}
\frac{\Lambda_k}{\Lambda_\phi} \simeq \frac{\displaystyle \frac{1}{N'}\sum\limits_{n=1}^{N'}\hat{\rho}_k(x'_{n})\alpha(x'_{n})}{\displaystyle \frac{1}{N_k}\sum\limits_{n=1}^{N_k}\phi(x_{n})\alpha(x_{n})}, \end{align}
where $\{x_n\}_{n=1}^{N_k}$ are samples from the local distribution $\hat{\rho}_k$, obtained using the neural network generator, and $\{x'_n\}_{n=1}^{N'}$ are samples from proposal distribution $\phi$. The estimator \eqref{eq:bridge_2} of ${\Lambda_k}/{\Lambda_\phi}$ depends on $\alpha$ and different choices of $\alpha$ lead to different statistical efficiency. Popular choices for $\alpha$ include $\alpha = 1/\sqrt{\hat{\rho}_k\phi}$, $\alpha = 1/\phi$ and $\alpha =  [\hat{\rho}_k^{1/t} + (A\phi)^{1/t}]^{-t}$ for pre-selected constants $t>0$ and $A>0$. In \cite{meng1996simulating}, the optimal $\alpha$ is shown to satisfy   
\begin{align}
\label{eq:alpha}
\alpha = \frac{N_k + N'}{N_k \hat{\rho}_k + \frac{\Lambda_k}{\Lambda_\phi} N'\phi}. 
\end{align}
Determining $\alpha$ from this relation requires an iterative process since it depends on the unknown quantity $\Lambda_k/\Lambda_\phi$. Starting with an initial guess of $\alpha$, $\Lambda_k/\Lambda_\pi$ is estimated using \eqref{eq:bridge_2}, then this new value of $\Lambda_k/\Lambda_\pi$ is used to update $\alpha$ via \eqref{eq:alpha}, iterating until convergence. We adopt this iterative approach in this work and direct the interested reader to \cite{meng1996simulating} for a detailed description. 

\subsection{Generating new samples using the trained generator}\label{sec:gen_sample}
After training the neural networks for all unimodal components and estimating the mixing ratio, we can assemble the final generator for the target multimodal distribution $\rho(x)$. Specifically, the generator 
$X = F(Y; \theta)$ in Eq.~\eqref{eq:transport} can be defined by
\begin{equation}\label{eq:final}
    F(Y, \theta) := \sum_{k=1}^K \mathbf{1}(\lambda = k)\; T_k(Y, \theta_k), 
\end{equation}
where $T_k(Y,\theta_k)$ is the transport map for the unimodal component $\hat{\rho}_k$, $\lambda$ is the random variable following the discrete probability distribution defined by the mixing ratio $\{\hat{r}_k\}_{k=1}^K$, i.e., 
\begin{equation}\label{eq:lambda}
    P(\lambda = k) = \hat{r}_k \;\; \text{ with }\;\; \hat{r}_k \ge 0 \;\text{ and }\; 
    \sum_{k=1}^K \hat{r}_k = 1.
\end{equation}
To generate one new sample, we first generate one sample from the discrete probability distribution in Eq.~\eqref{eq:lambda} to pick which unimodal component will be sampled, the we can draw a random sample of $Y$ from the standard Gaussian distribution and push the sample through the $k$-th generator $T_k(Y,\theta_k)$ to generate one sample of the target distribution $\rho(x)$. For a large number of samples, the allocation of the samples to different modes will be controlled by estimated mixing ratio via the probability distribution of $\lambda$. 

\subsection{Discussion on the comparison between our method traditional sampling methods}\label{sec:discussion}
Our method has several appealing advantages over traditional sampling approaches, such as MCMC and Langevin dynamics, particularly in handling high-dimensional multimodal distributions. One of the primary challenges in conventional MCMC-based methods is the inefficiency of mode-hopping, where chains can become trapped in a single mode for long periods, leading to slow mixing and biased sampling. In contrast, our approach leverages a divide-and-conquer strategy, where we construct a dedicated generator for each unimodal component separately. This decomposition allows us to model complex multimodal distributions without relying on local proposals or requiring difficult inter-mode transitions. Instead of depending on slow, sequential updates as in MCMC, we estimate the correct mixing ratio using Gaussian bridge sampling, which efficiently reconstructs the overall multimodal structure by integrating the separately trained unimodal generators. This allows our method to bypass energy barriers, enabling efficient and accurate sampling of multimodal distributions. Furthermore, once the generative model is trained, it can generate independent samples in parallel, unlike MCMC, which requires long chains and burn-in phases to reach equilibrium. This parallelization is particularly advantageous in high-dimensional spaces where MCMC struggles with slow convergence and requires extensive computational resources. By avoiding these bottlenecks, our method provides a scalable, robust, and efficient alternative for sampling from complex, multimodal distributions.


\section{Numerical experiments}
\label{sec:num}
In this section, we demonstrate and evaluate the performance of our framework using five examples. First, to illustrate its internal workings, we conduct a simple test involving a $2d$ Gaussian mixture model comprising two Gaussian distributions. By varying the distance between the modes, we explore scenarios where they are well-separated, weakly connected, or completely overlapping, and show that our framework performs effectively in all three cases. The second example demonstrates the applicability of our method in high-dimensional problems, using a 100$d$ mixture model with two Gaussian distributions. Then, we evaluate our framework in scenarios where local modes exhibit more complex and asymmetric shapes. In the third example, we introduce skewness to the modes by employing a 20$d$ mixture model consisting of four skew-normal distributions. To further showcase the versatility of our approach, the fourth example isolates and evaluates the core module of our framework focused on single component sampling (Steps 2-4), where the target densities display extremely intricate patterns, derived from 2$d$ images. Lastly, we apply our method to solve an inverse partial differential equation (PDE) problem modeling the contamination flows in the water. The performance of our framework is evaluated and compared against other popular sampling methods, highlighting its efficiency.

\subsection{$2$d Gaussian mixture model}
\label{sec:test1}
\begin{figure}[h]
\centering
\includegraphics[width = 0.8\textwidth]{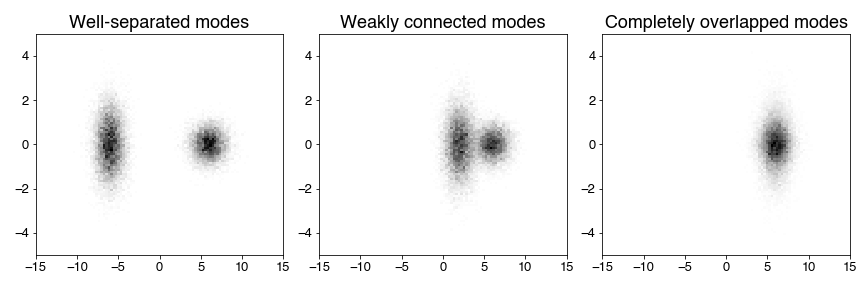}
\vspace{-.1in}
\caption{Mixture models consisting of two Gaussian modes at different distances, defined as in \eqref{eq:GM_mixture_2d}. Note that the $x$-axis and $y$-axis are scaled differently.}
\label{fig:GM_mixture}
\end{figure}

We consider the Gaussian mixture model 
\begin{equation}
    \rho(x) = r_1 \mathcal{N}(x \mid \mu_1, \Sigma_1) + r_2 \mathcal{N}(x \mid \mu_2, \Sigma_2),
\end{equation}
where \( \mathcal{N}(x \mid \mu, \Sigma) \) represents the multivariate normal distribution with mean $\mu\in \mathbb{R}^2$ and covariance matrix $\Sigma\in \mathbb{R}^{2 \times 2}$. The two modes of the distribution are specified as
\begin{gather}
\begin{aligned}
\mu_1 &= (6,0) ,\ \Sigma_1 = \begin{bmatrix} 1.2 & 0 \\ 0 & 0.5 \end{bmatrix}, \\
\mu_2 &= (a,0) ,\ \Sigma_2 = \begin{bmatrix} 1 & 0 \\ 0 & 1 \end{bmatrix}. \end{aligned}
\label{eq:GM_mixture_2d}
\end{gather}
We consider three different values for $a$, namely $a = -6,\, a=2,\, a=6$, corresponding to three settings: i) well-separated modes, ii) weakly connected modes, and iii) completely overlapping modes. The mixing coefficients are set to $r_1 = 0.4,\, r_2=0.6$. An illustration of these tested mixture models are shown in Figure \ref{fig:GM_mixture}. 

\begin{figure}[h]
\centering
\includegraphics[width = 0.3\textwidth]{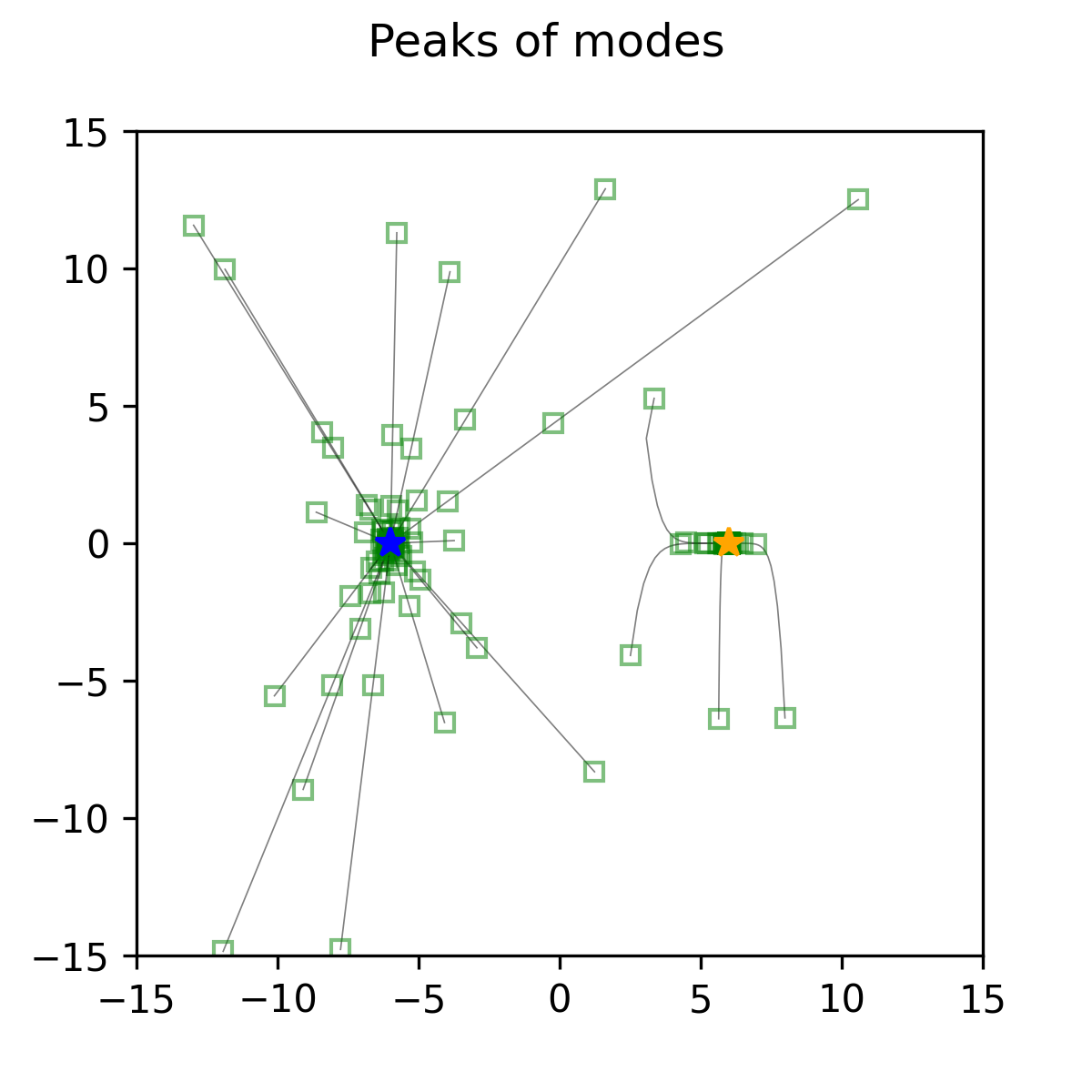}
\includegraphics[width = 0.3\textwidth]{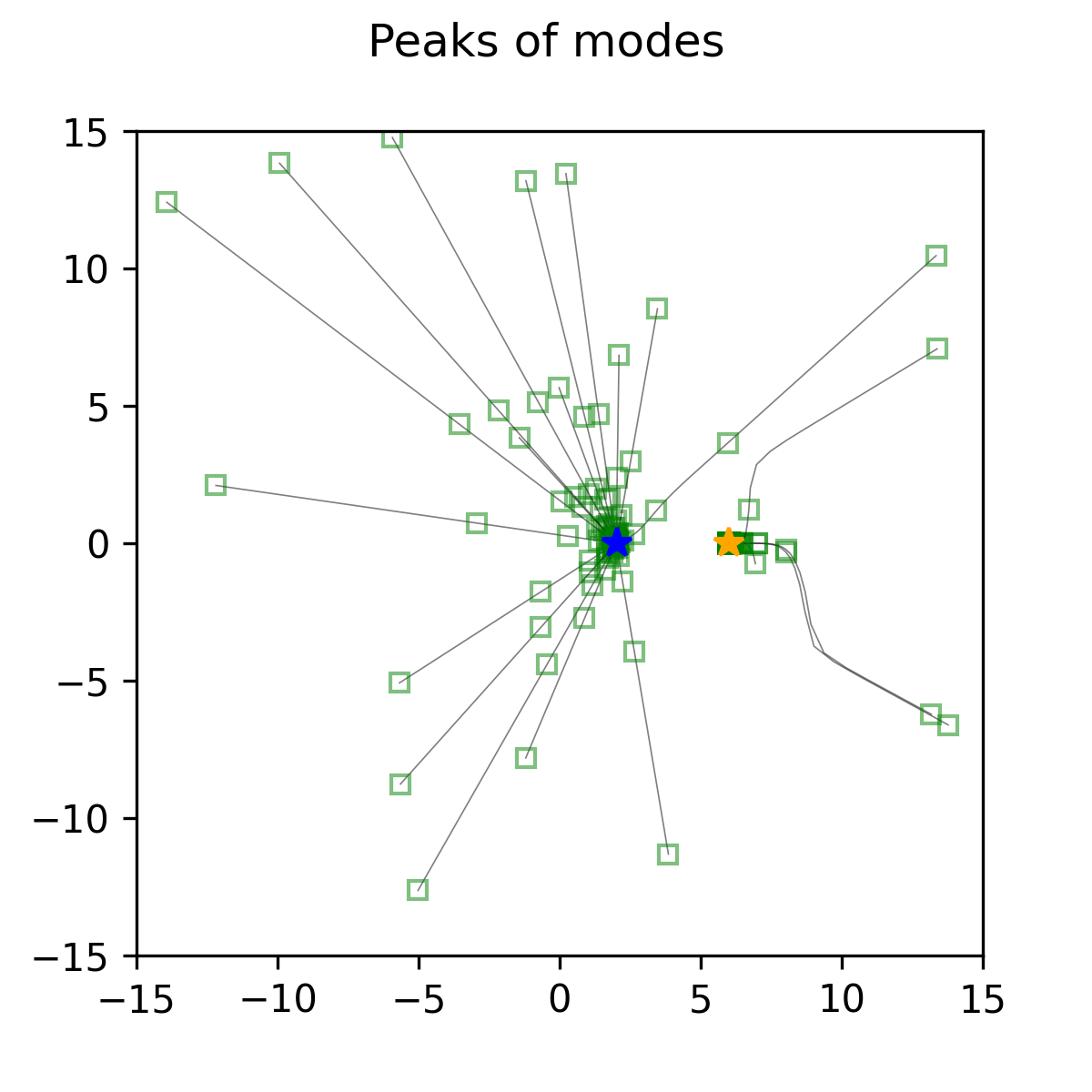}
\includegraphics[width = 0.3\textwidth]{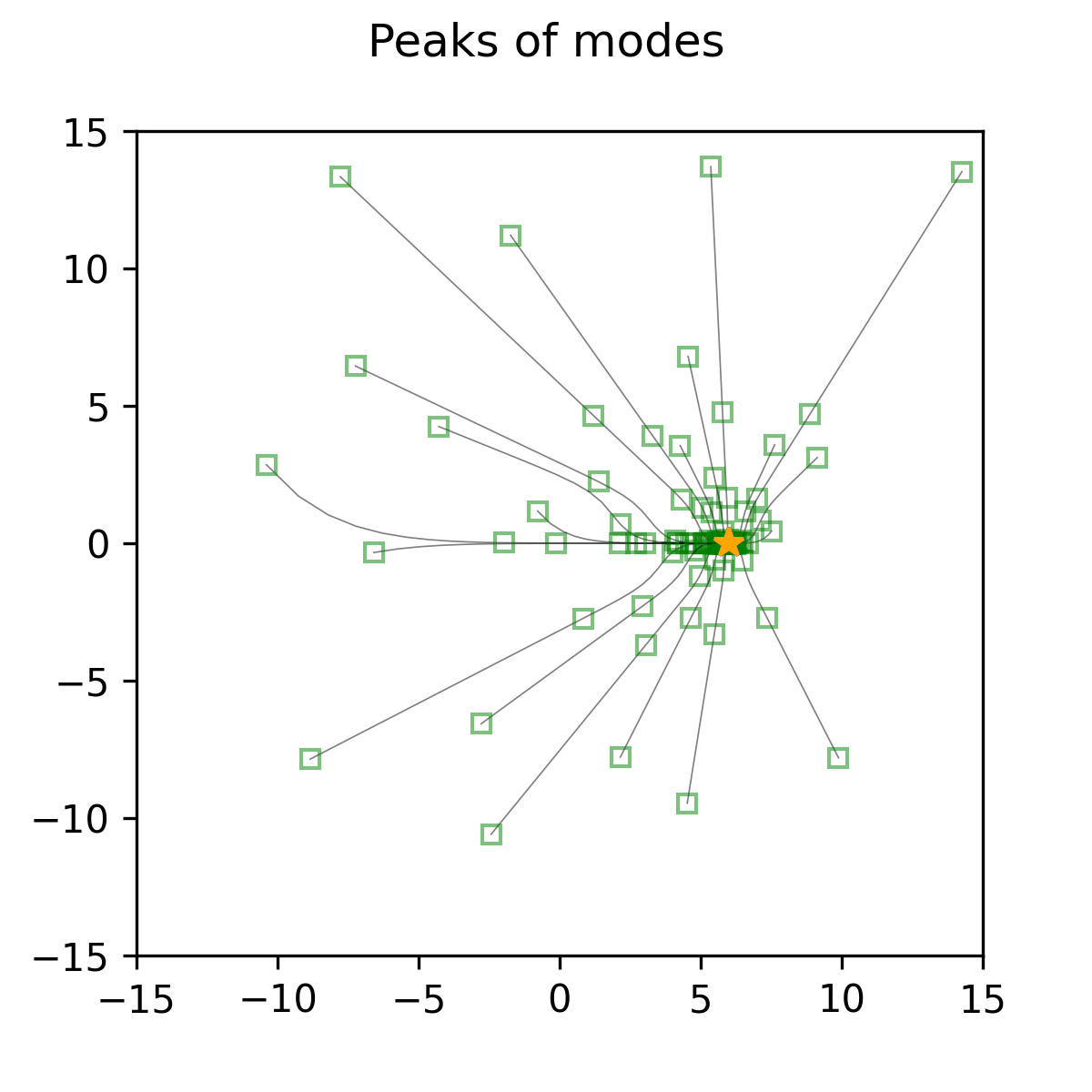}
\vspace{-.1in}
\caption{Peaks of target PDF modes found by multi-start gradient descent. From left to right: well separated modes, weakly connected modes, and completely overlapped modes. }
\label{fig:2D_SGD}
\end{figure}

In \textbf{Step 1}, we perform a multi-start optimization with gradient descent (see \eqref{eq:GD}) to identify the peaks of the modes. We initialize the optimization with $2000$ starting points uniformly sampled within the box $[-15,15]^2$, which covers the support of target distribution. Each run consists of $2000$ iterations with a fixed step size $\lambda=0.1$. The optimization paths are shown in Figure \ref{fig:2D_SGD}. We observed that for the first two cases (well-separated and weakly connected modes), the algorithm successfully identifies two distinct optima, while in the last case, a single optimum is found since the peaks of two modes coincide. 

\begin{figure}[h]
\centering
\includegraphics[width = 0.35\textwidth]{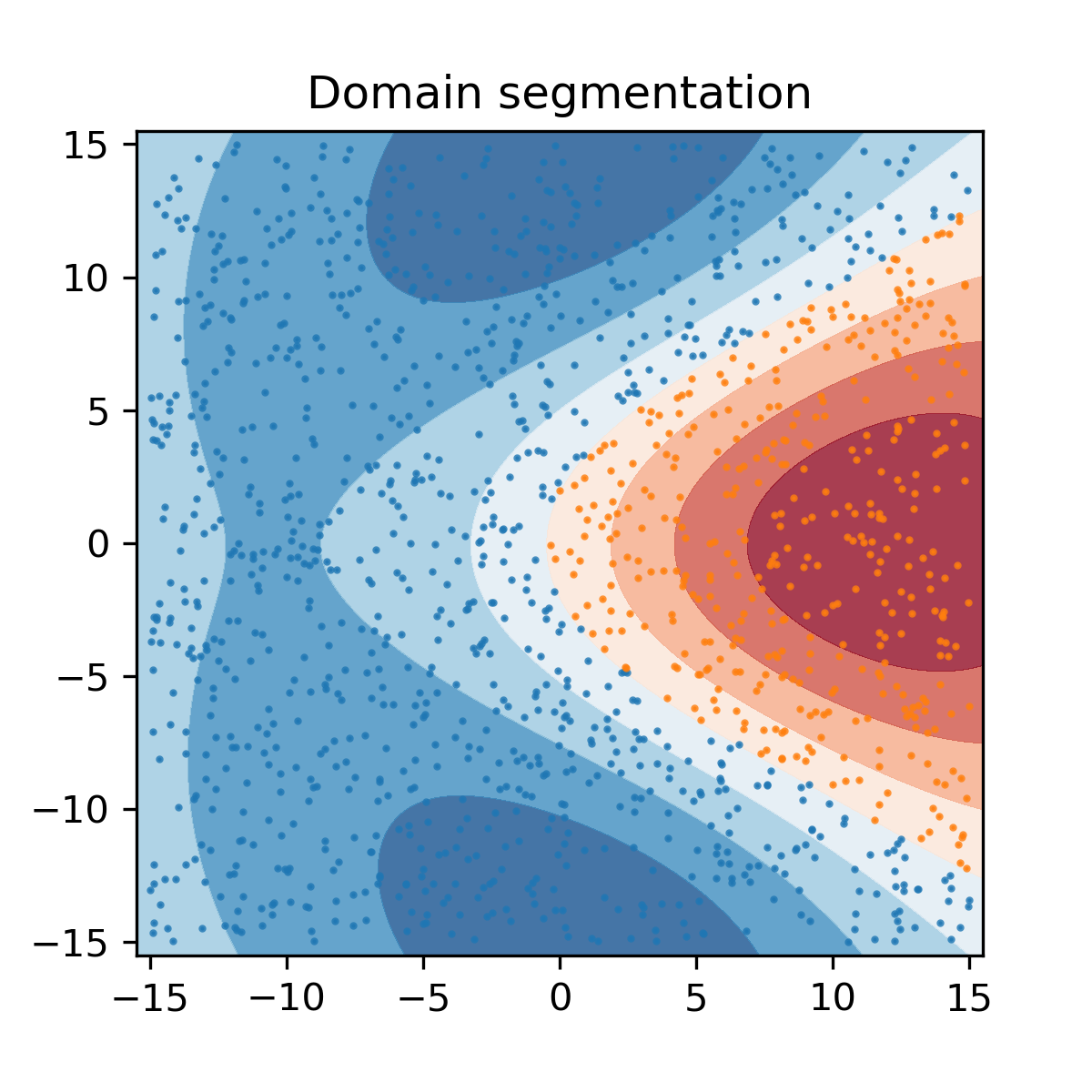}
\includegraphics[width = 0.35\textwidth]{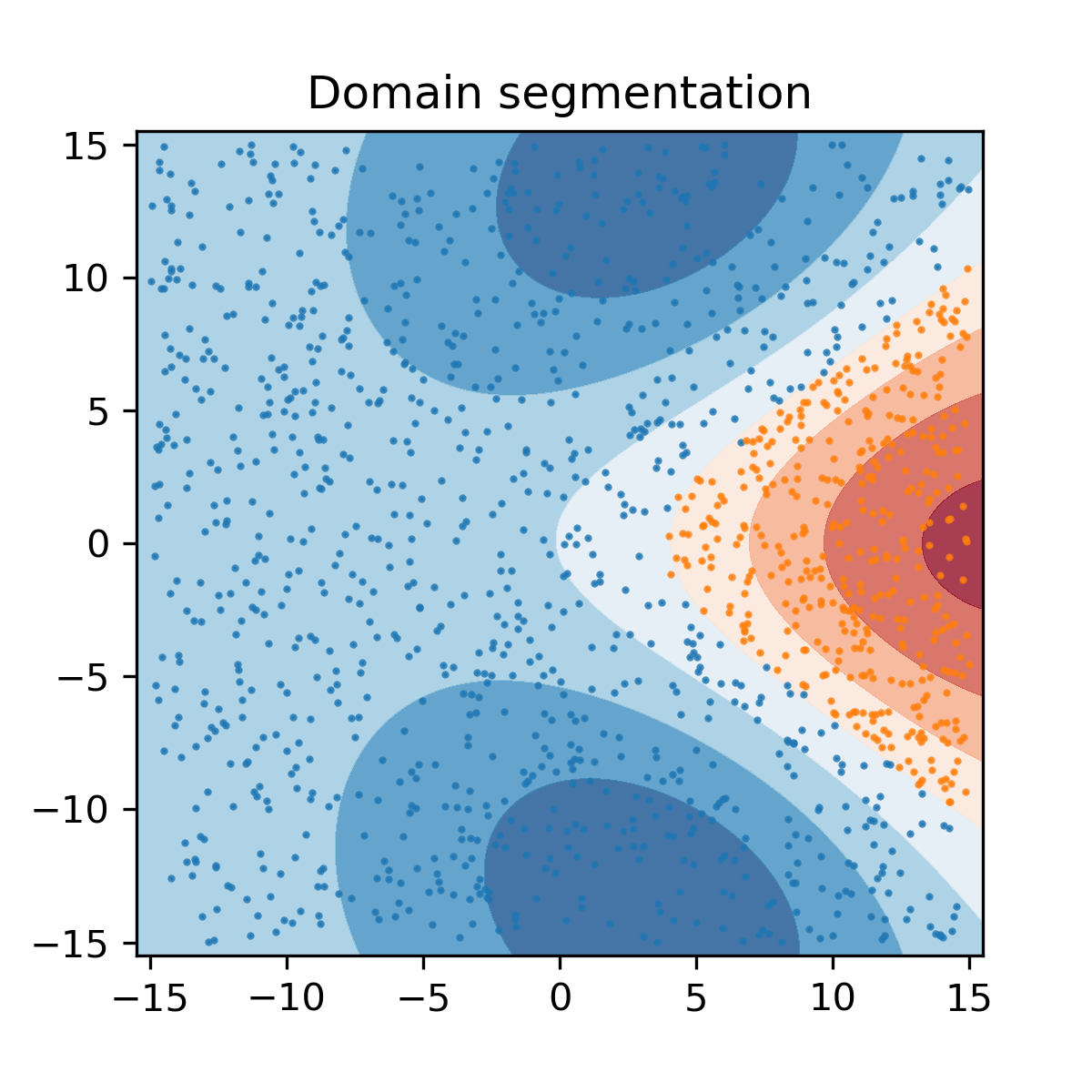}
\vspace{-.1in}
\caption{The segmentation of PDF domain into separate subdomains: (left) well-separated modes ($a=-6$); (right) weakly connected modes ($a=2$). The case of completely overlapped modes ($a=6$) is omitted here since no segmentation is necessary. The shaded colors represent the decision function values, with each region (reddish vs. bluish) corresponding to a class. Darker regions indicate areas where the classifier is more confident in its classification, and the curve where colors change from blue to red represents the decision boundary.}
\label{fig:2D_CSVC}
\end{figure}

Next, we conduct the segmentation of the probability space according to the support of each mode. This step is required only for the first two scenarios, as in the final case, the modes coincide and the target distribution is actually unimodal. We begin by labeling the initial points based on the optima they converge to during the optimization phase. Then, the C-SVC algorithm is applied to delineate the decision boundaries. The resulting segmentation of the probability domain is shown in Figure \ref{fig:2D_CSVC}. We note that the segmentation does not require the supports of the local components to be completely disjoint. In cases where the supports overlap (as in the second scenario), the shared region will be partitioned and assigned to one of the modes.

\begin{figure}[h]
\centering
\includegraphics[width = 0.99\textwidth]{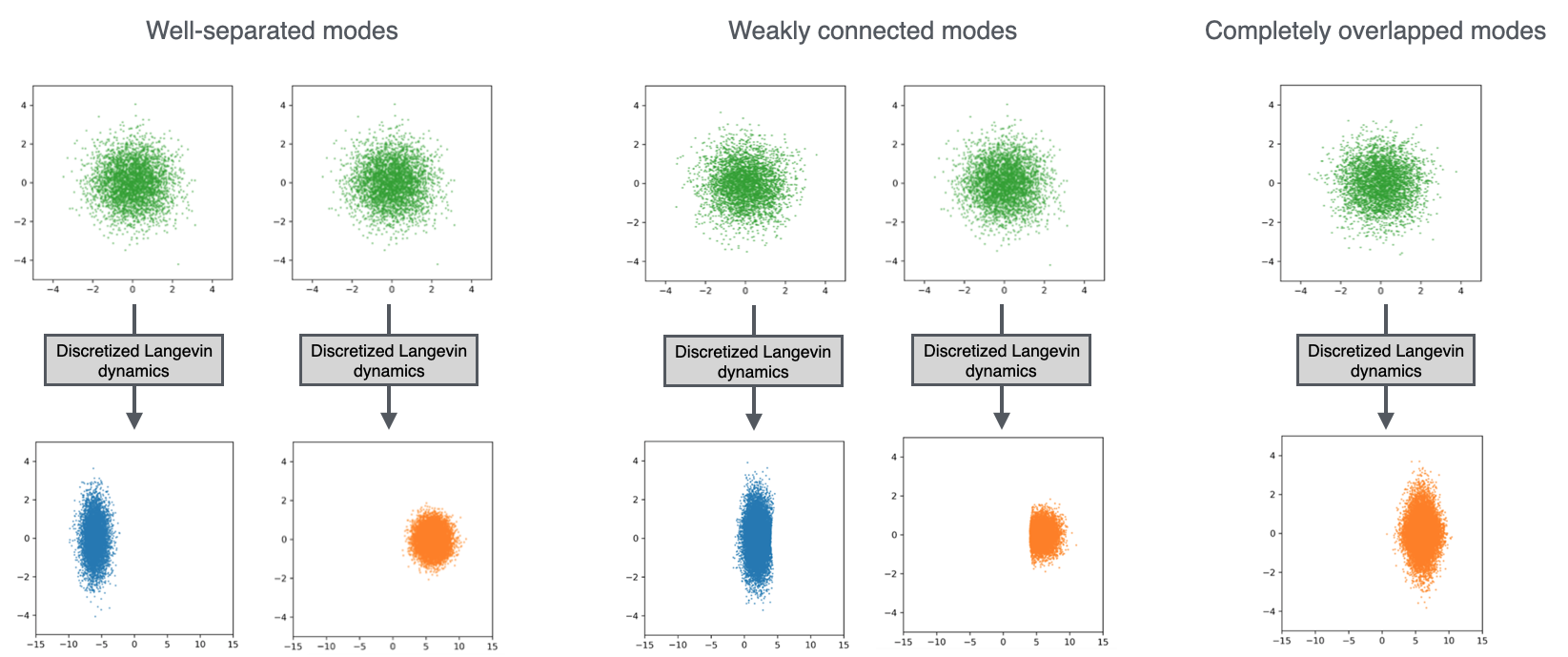}
\vspace{-.1in}
\caption{The generation of samples for target unimodal components using Langevin dynamics. Each unimodal component is sampled independently. Top: Input samples. Bottom: Output samples obtained from the Langevin scheme \eqref{eq:lgv}.}
\label{fig:lgv}
\end{figure}

The target distribution restricted within each subregion becomes unimodal. In the next steps, we develop a generative model for each unimodal component, which is a significantly easier task compared to directly sampling from the original multimodal distributions. In \textbf{Step 2}, using the discretized Langevin dynamics \eqref{eq:lgv}, we transform an initial set of normally distributed points into samples from the desired target unimodal components. Figure \ref{fig:lgv} plots the initial sample points together with the output samples obtained after running the Langevin dynamics. We independently sample two unimodal components in cases of well-separated modes and weakly connected modes, while for the overlapped modes setting, we sample a single unimodal component which is also the target distribution. For each component, the Langevin scheme is conducted using $10000$ samples and step size $\eta = 0.001$.   

\begin{figure}[!h]
\centering
\includegraphics[width = 0.25\textwidth]{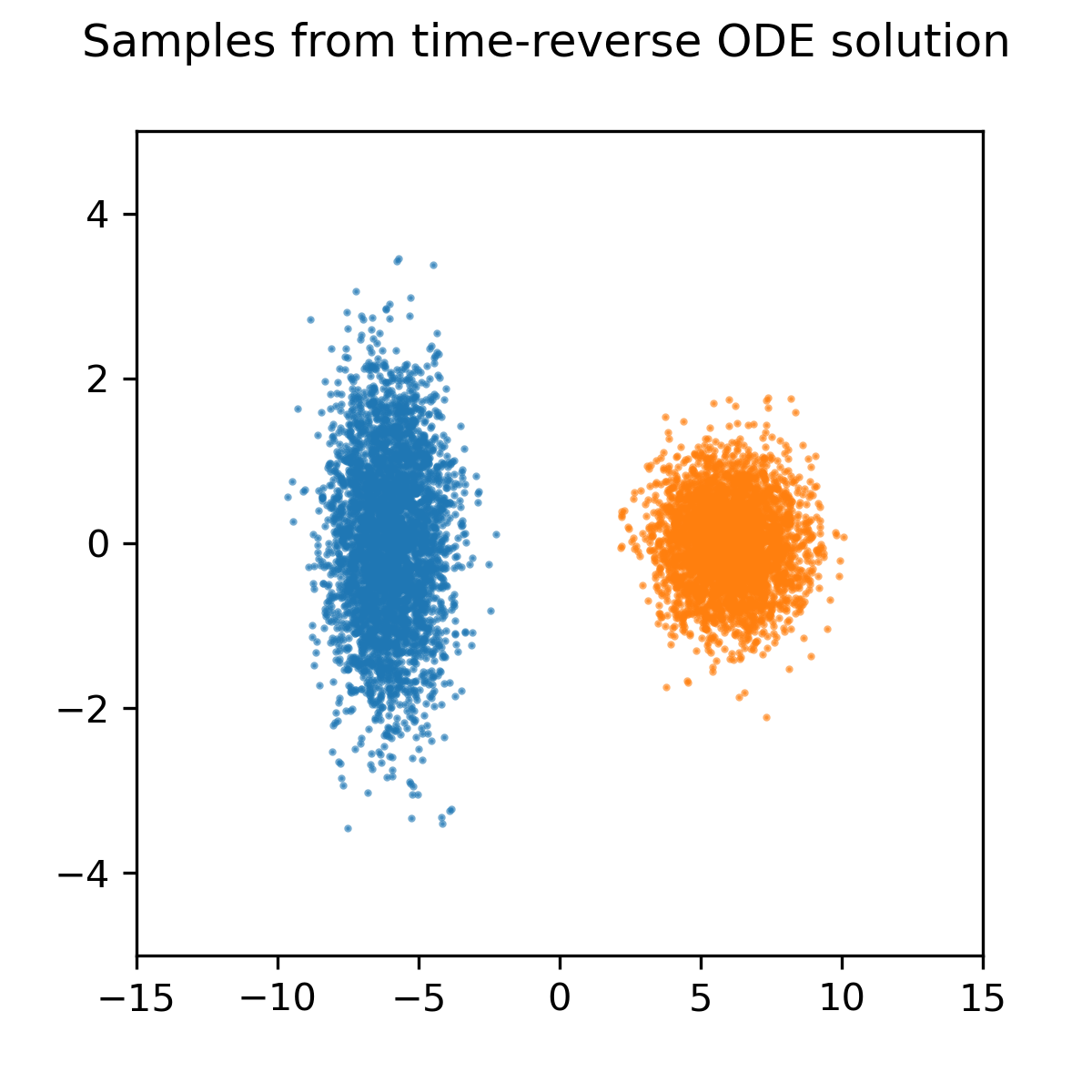}
\includegraphics[width = 0.25\textwidth]{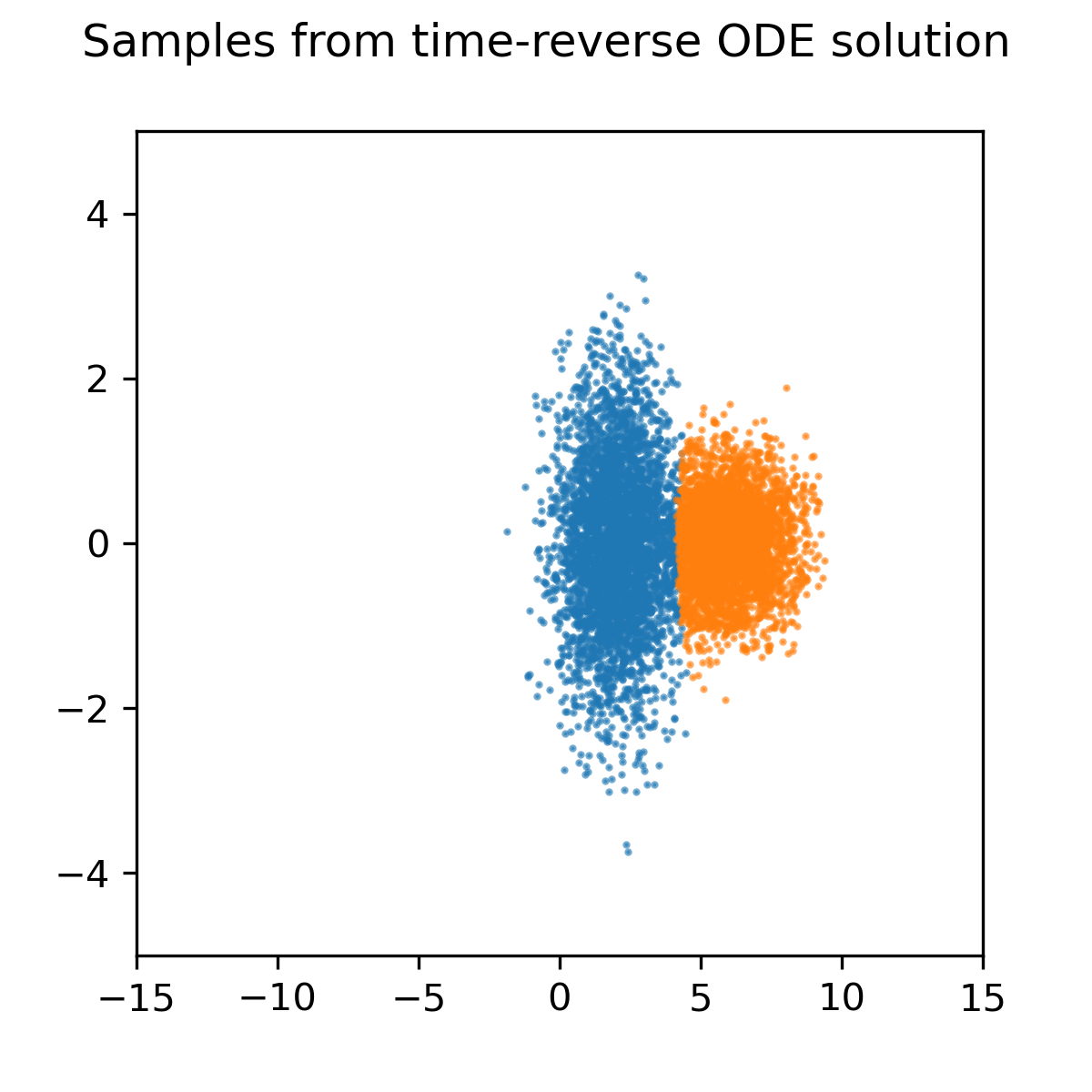}
\includegraphics[width = 0.25\textwidth]{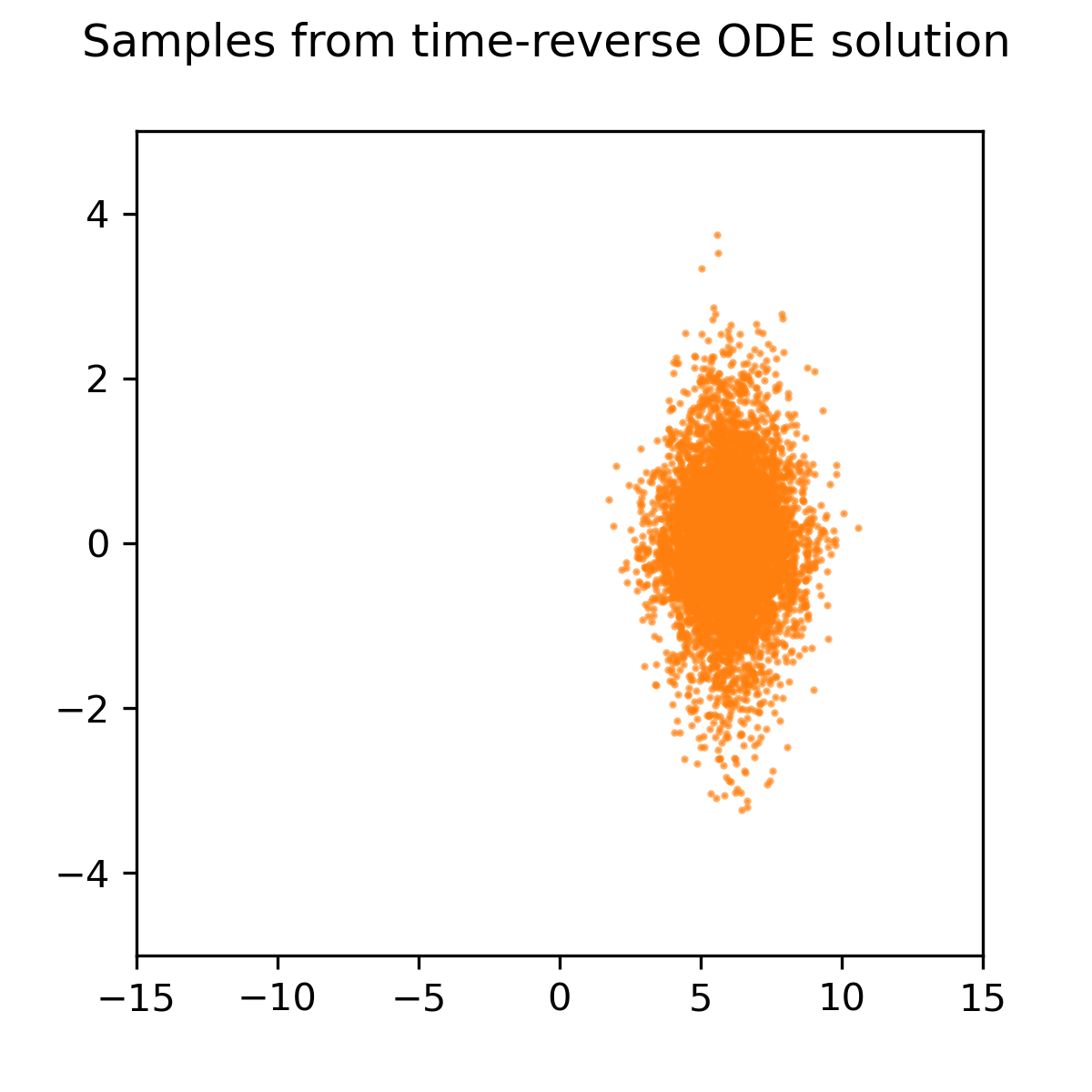}
\vspace{-.1in}
\caption{The output labeled data generated by solving the time-reverse ODEs \eqref{DM:RSDE}. Samples for each unimodal component, distinguished here by different colors, are generated independently from different diffusion models. }
\label{fig:diff}
\end{figure}

In \textbf{Step 3}, we generate labeled data for each component by constructing a map that corresponds each Gaussian sample with a corresponding sample from the target component. To construct this map, we employ the score-based diffusion model described in Section \ref{sec:diff} and solve the reverse ODE \eqref{DM:RSDE} where the score function is estimated by Monte-Carlo method with samples generated from the Langevin dynamics in Step 2. For each component, we discretize the time interval into 100 steps and solve the reverse ODE with $10000$ Gaussian samples from the terminal state. The resulting outputs are $10000$ samples from the initial states (i.e., target unimodal components), which are shown in Figure \ref{fig:diff}.

\begin{figure}[!h]
\centering
\includegraphics[width = 1.0\textwidth]{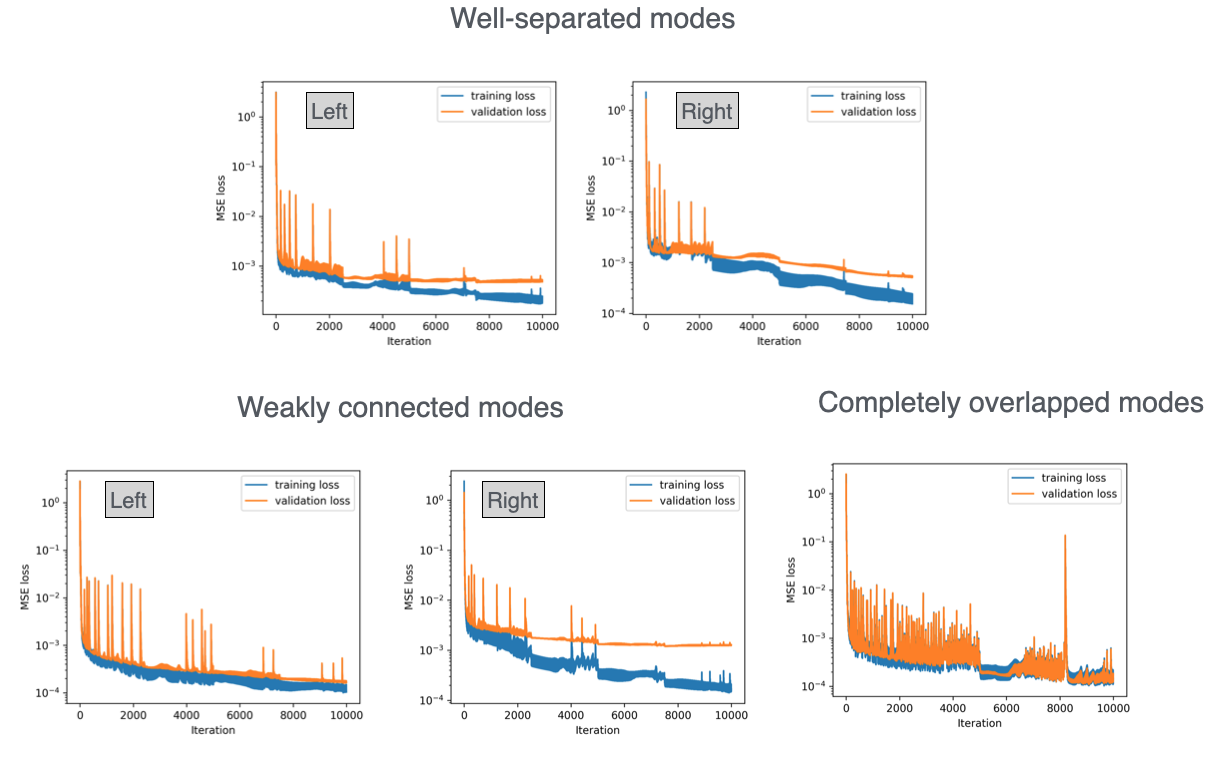}
\vspace{-.1in}
\caption{The evolution of training and validation errors in supervised learning of the neural-network-based samplers for the unimodal components of Gaussian mixture models. From left to right: the left and right components in the well-separated model, the left and right components in the weakly connected model, and the single component in the fully overlapped model. }
\label{fig:fnn}
\end{figure}

In \textbf{Step 4}, we train neural network models to generate samples for each unimodal component from randomly drawn Gaussian samples. Using the labeled data produced by our diffusion models, we adopt a supervised learning approach with a relatively simple neural network architecture: a feedforward model consisting of three hidden layers, each with $1000$ nodes and tanh activation functions. Separate neural networks are trained for each unimodal target component. The training process employs the Adam optimizer with PyTorch's default parameters, except for the learning rate, which starts at $0.001$ and is halved every $500$ epochs. The training is conducted for $2000$ epochs with a batch size $1500$. As our dataset contains $10000$ input-output pairs for each unimodal component, we use $8000$ pairs for training and $2000$ for validation. The evolutions of training and validation errors for each component are presented in Figure \ref{fig:fnn}. We observe that in all cases, the training errors drop fast at the beginning, reaching the range of $10^{-3}$, and continues to improve gradually with additional epochs. By the end of the training, the errors reach approximately $10^{-4}$. The model generalizes well in most cases, as the validation errors remain close to the training errors. We note that it is not required for the models to match individual points exactly; rather, it suffices that the generated samples follow the correct target distribution. 

\begin{figure}[!h]
\centering
\includegraphics[width = 0.75\textwidth]{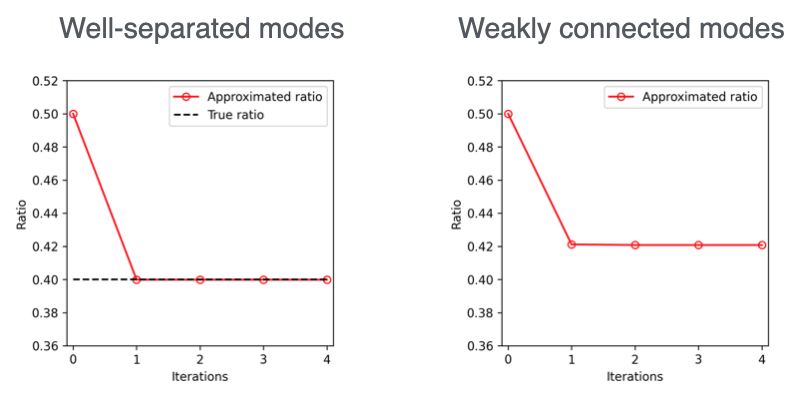}
\vspace{-.1in}
\caption{The evolution of the predicted mixing ratio $\hat{r}_1$ associated with the right mode in case of well-separated modes (left) and weakly connected modes (right). Equilibrium is achieved after just one iteration of the Gaussian bridge sampling iterative scheme. In case of well-separated modes where the true value of mixing ratio is known, the predicted mixing ratio reach the true value.}
\label{fig:ratio}
\end{figure}

Since the generative model for each unimodal component is developed independently without communicating to one another, in \textbf{Step 5}, we estimate the weights of the unimodal components of the target distribution to correctly attribute samples to each mode. This step only involves in the well-separated and weakly connected scenarios where the target models have two distinct modes. Here, we employ the Gaussian bridge sampling approach with the iterative process outlined in Section \ref{sec:bridge_sampling} to compute the normalizing constant $\Lambda_k$ of each $\hat{\rho}_k$, and through that, the associated mixing ratio $\hat{r}_k$. In Figure \ref{fig:ratio}, we show the evolution of the mixing ratio $\hat{r}_1$ of the right mode, where equilibrium is achieved after just one iteration for both well-separated and weakly connected scenarios. The mixing ratio $\hat{r}_2$ of the left mode can be trivially computed as $\hat{r}_2= 1 - \hat{r}_1$ and thus is not shown here. In case of well-separated modes, we verified that $\hat{r}_1 = r_1$, while in case of weakly connected modes, a ground truth for $\hat{r}_1$ does not exists because it depends on the partition of the joint support of the modes by C-SVC. Nevertheless, we can evaluate the quality of the mixing ratios by testifying that new samples generated by the final generator follow the target mixture distribution.

\begin{figure}[!h]
\centering
\includegraphics[width = 0.9\textwidth]{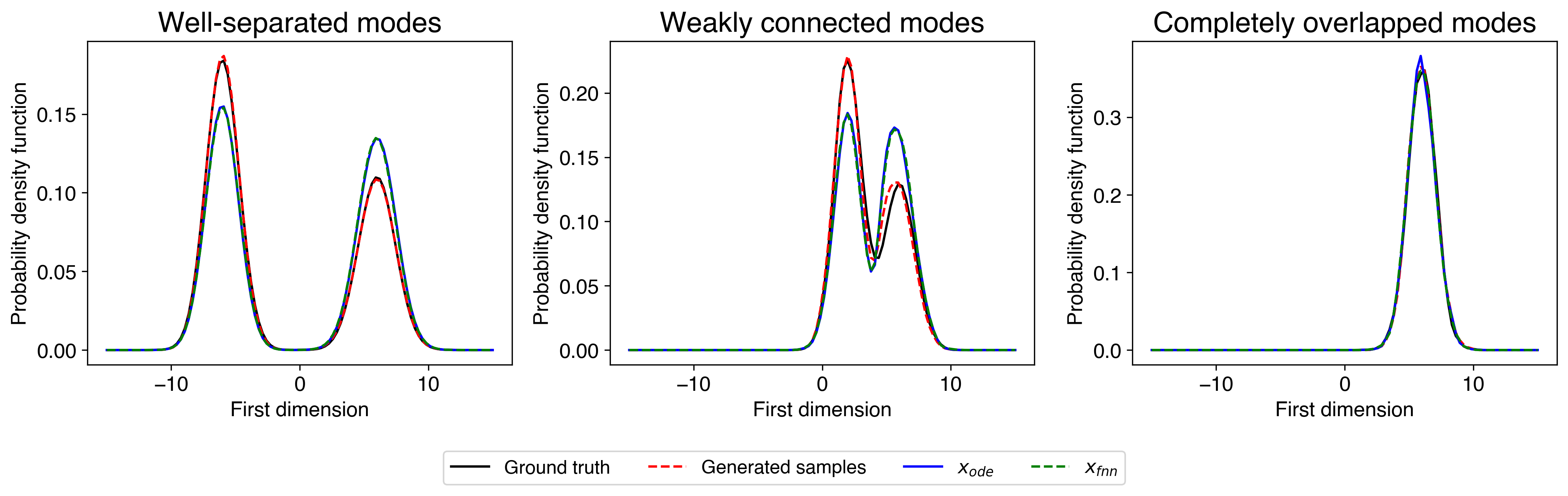}
\vspace{-.1in}
\caption{The marginal distribution of samples generated from the final assembled generators ({`Generated samples'}) in the first dimension compared to the ground truth ({`Ground truth'}). Also included in the plots are the samples generated in the intermediate steps of our workflow: labeled samples obtained from solving the reverse ODEs in Step 3 (`$x_{ode}$') and samples generated by naively combining independent neural network models trained in Step 4 (`$x_{fnn}$').}
\label{fig:test1_pdf1D}
\end{figure}

 Now, we have trained all neural network models for unimodal components and obtained the mixing ratios. In \textbf{Step 6}, we generate $20000$ new samples for the original target distribution using the final assembled generator. Figure \ref{fig:test1_pdf1D} presents the marginal distribution of these samples in the first dimension. We compare them against the `Ground truth' samples generated from the built-in \texttt{multivariate\_normal} function from the \texttt{scipy.stats} package. Additionally, samples generated in the intermediate steps of our workflow are also illustrated. In particular, we plot the marginal distribution of the labeled samples obtained in Step 3 from solving the reverse ODEs (`$x_{ode}$'), and that of the samples generated in Step 4 by independent neural network models for each unimodal component (`$x_{fnn}$'). Several observation can be drawn from the plots. First, the PDFs of final generated samples coincide with the ground truth, verifying the efficiency of our framework. Second, the PDFs of $x_{ode}$ samples and $x_{fnn}$ samples are also closely aligned, demonstrating that the neural network models trained on labeled data can effectively sample the target components, despite the possible presence of a generalization gap in point-to-point matching (see Figure \ref{fig:fnn}). Third, samples from reverse ODEs and neural network generators successfully capture unimodal distributions, as shown in Figure \ref{fig:test1_pdf1D} (right). When the target distribution has multiple modes, there is a notable discrepancy between the samples from the combined neural network models before and after the mixing ratios are estimated and applied. Here, the PDFs of $x_{ode}$ and $x_{fnn}$ in the well-separated and weakly connected cases reflect our initial blind assignment of $10000$ samples to each unimodal component. The fact that the final assembled generators accurately sample from the ground truth validates the accuracy of our estimated mixing ratios.

\subsection{$100$d Gaussian mixture model}
\label{sec:test2_gaussian}
In this second test, we consider the $100d$ Gaussian mixture model 
\begin{equation}
    \rho(x) = r_1 \mathcal{N}(x \mid \mu_1, \Sigma_1) + r_2 \mathcal{N}(x \mid \mu_2, \Sigma_2),
    \label{eq:100dGaussian_1}
\end{equation}
where \( \mathcal{N}(\cdot \mid \mu, \Sigma) \) represents the multivariate normal distribution with mean $\mu\in \mathbb{R}^{100}$ and covariance matrix $\Sigma\in \mathbb{R}^{{100} \times {100}}$. The two modes of the distribution are specified as
\begin{gather}
    \label{eq:100dGaussian_2}
\begin{aligned}
\mu_1 &= (6,0,\ldots,0) ,\ \Sigma_1 = \begin{bmatrix} \hat{\Sigma}_1 & 0 \\ 0 & I_{96} \end{bmatrix},\ \text{with }\hat{\Sigma}_1 =  \begin{bmatrix} 1.2 & 0 & 0 & 0 \\ 0 & 0.8 & 0 & 0 \\ 0 & 0 & 1 & 0 \\ 0 & 0 & 0 & 0.5 \end{bmatrix}, \\
\mu_2 &= (a,0,\ldots,0) ,\ \Sigma_2 = I_{100}. 
\end{aligned}
\end{gather}
where $I_d$ represents the identity matrix of size $d\times d$. Let $\Delta$ denote the distance between two modes. Similar to Test 1, we consider three different values for $a$, namely $a = -6,\, a=2,\, a=6$, corresponding to well-separated modes ($\Delta = 12$), weakly connected modes ($\Delta = 4$), and completely overlapping modes ($\Delta = 0$) scenarios. The mixing coefficients are set to $r_1 = 0.6,\, r_2=0.4$. 

We evaluate the performance of our framework against two traditional probabilistic sampling methods: (i) Emcee \cite{goodman2010ensemble,Emcee}, an ensemble MCMC method that utilizes multiple walkers exploring the parameter space in parallel, and (ii) No-U-Turn Sampler (NUTS) \cite{10.5555/2627435.2638586}, an adaptive variant of Hamiltonian Monte Carlo that automatically tunes step sizes and trajectory lengths. The experimental setup of our methodology is detailed in Table \ref{tab:hyperparameters}. To demonstrate that our framework does not require excessive fine-tuning of hyperparameters across different problems, we use a similar setup to the $2d$ Gaussian mixture test (see Section \ref{sec:test1}), with the main difference being that in the first step, we let the SGD run longer for mode identification. In Figure \ref{fig:test2_ratio}, we show the convergence of the mixing ratio $\hat{r}_2$ of the left mode, estimated using the Gaussian bridge sampling iterative scheme in the well-separated case. Here, the true value is obtained after just five iterations. The final trained generator is used to generate $20000$ samples of the target distribution for evaluation. 

For Emcee, we use a Python implementation from \cite{Emcee}.  We employ $400$ walkers, following their recommendation that the number of walkers should be at least twice as the model dimension. Normal distribution with zero mean and different standard deviations are tested for sampling the initial guess for each walker. The best results, which are reported here, correspond to a standard deviation of $5$. To further mitigate the impact of poor initialization, we allow a long burn-in period of $2000$ steps, followed by $500$ sampling steps, resulting in a total of $400 \times 500 = 200000$ collected samples. For NUTS, we use the publicly available Python package from \cite{10.5555/2627435.2638586}. Both the number of burn-in steps and the number of sampling steps are set to $20000$, with initial guesses drawn from a standard normal distribution. We test different values for target mean acceptance probability $\delta$, ranging from $0.2$ to $0.8$, all yielding similar results. However, higher $\delta$ increases the computational runtime. Here, we present the results with $\delta =0.2$.  

\vspace{.1in}
\begin{minipage}{0.55\textwidth}
   \centering
    \resizebox{1.0\textwidth}{!}{  
    \renewcommand{\arraystretch}{1.2} 
    \begin{tabular}{lc}
        \toprule
        \textbf{Step} & \textbf{Setup} \\
        \midrule
        \multirow{2}{*}{Step 1}  & 2000 starting points within \( [-15,15]^{100} \) \\
                                 & 20000 SGD iterations with step size \( \lambda = 0.2 \) \\
        \midrule
        \multirow{2}{*}{Step 2}  & 10000 samples per component \\
                                 & 10000 Langevin iterations with step size \( \eta = 0.001 \) \\
        \midrule
         \multirow{2}{*}{Step 3} &   explicit Euler scheme for reverse ODE with 100 time steps \\
                                & 10000 labeled samples generated per component \\
        \midrule
        \multirow{4}{*}{Step 4}  & Feedforward model with three hidden layers \\
                                 & Each layer has 800 nodes with \texttt{tanh} activation \\
                                 & Adam with initial learning rate 0.001, halved every 500 epochs \\
                                 & Training over $1000$ epochs with batch size $1500$ \\
        \bottomrule
    \end{tabular}
    }
    \captionof{table}{Experimental setup of our framework for training a generative model for the $100d$ Gaussian mixture distribution \eqref{eq:100dGaussian_1}-\eqref{eq:100dGaussian_2}.}
    \label{tab:hyperparameters}
\end{minipage}
\hfill
\begin{minipage}{0.35\textwidth}
\centering
\centering
\includegraphics[width = 0.85\textwidth]{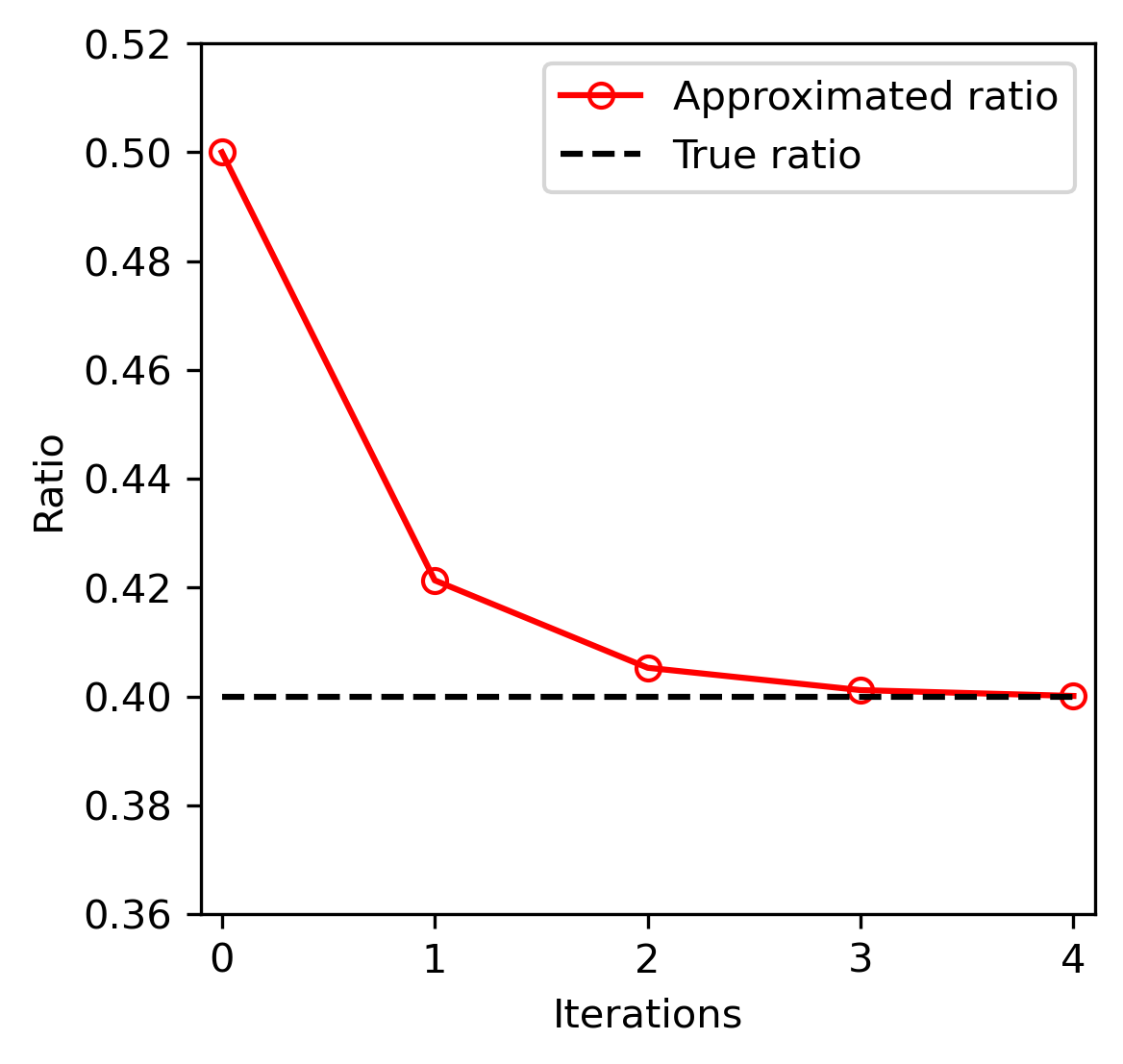}
\vspace{-.1in}
\captionof{figure}{The convergence of the predicted mixing ratio $\hat{r}_2$ associated with the left mode in case of well-separated modes.}
\label{fig:test2_ratio}
\end{minipage}
\vspace{.2in}

\begin{figure}[!h]
\centering
\includegraphics[width = 0.9\textwidth]{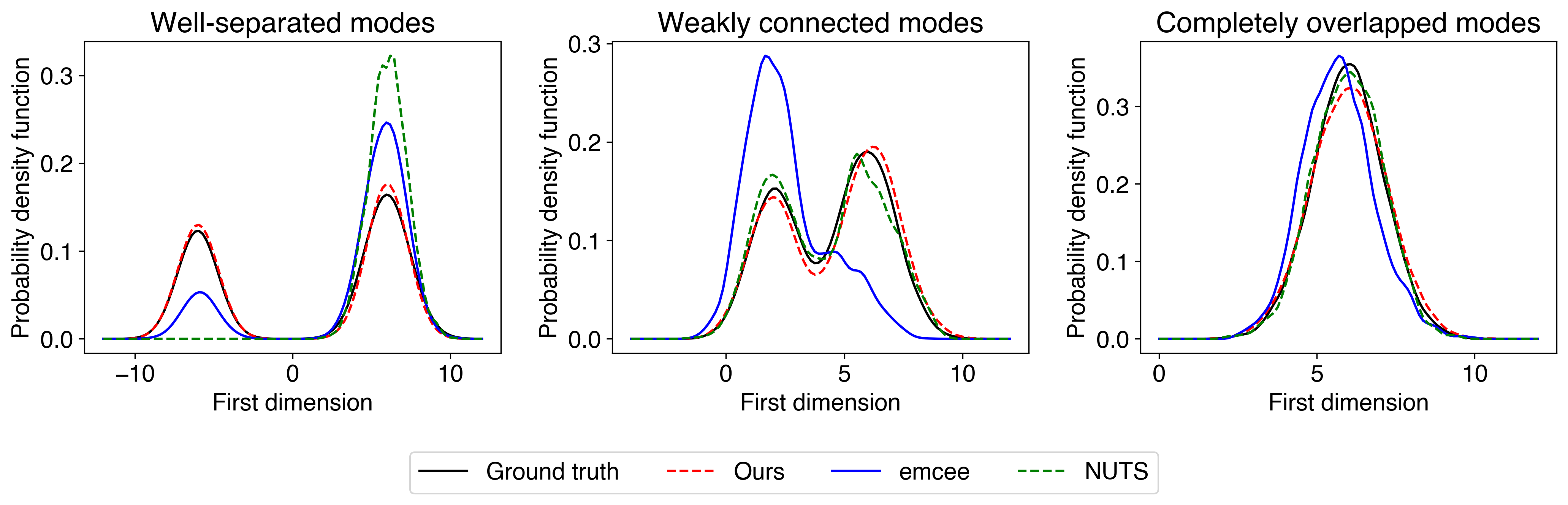}
\vspace{-.1in}
\caption{Marginal distribution (in the first dimension) for samples from our final generators versus the ground truth, Emcee, and NUTS. Our samples closely match the ground truth, while NUTS misses the left mode in the well-separated case and Emcee exhibits poor mixing in both well-separated and weakly connected scenarios.}
\label{fig:test2_pdf100D}
\end{figure}

In Figure \ref{fig:test2_pdf100D}, we show the marginal distribution of samples produced from our trained generator and the compared methods in the first dimension, where the mixture model exhibits bimodality. For reference, we also include ``ground truth'' samples, obtained using the built-in \texttt{multivariate\_normal} function from \texttt{scipy.stats} for each mode, with the correct mixture ratios applied. In all three cases, we observe that our generated samples accurately approximate the distribution of the ground truth, while NUTS completely misses the left mode in the well-separated model, and Emcee suffers from poor mixing in both well-separated and weakly connected cases. However, in the simplest scenario, where the modes are fully overlapping, NUTS takes a slight lead in providing the most accurate sampling of the target distribution.

\begin{figure}[!h]
\centering
\includegraphics[width = 0.9\textwidth]{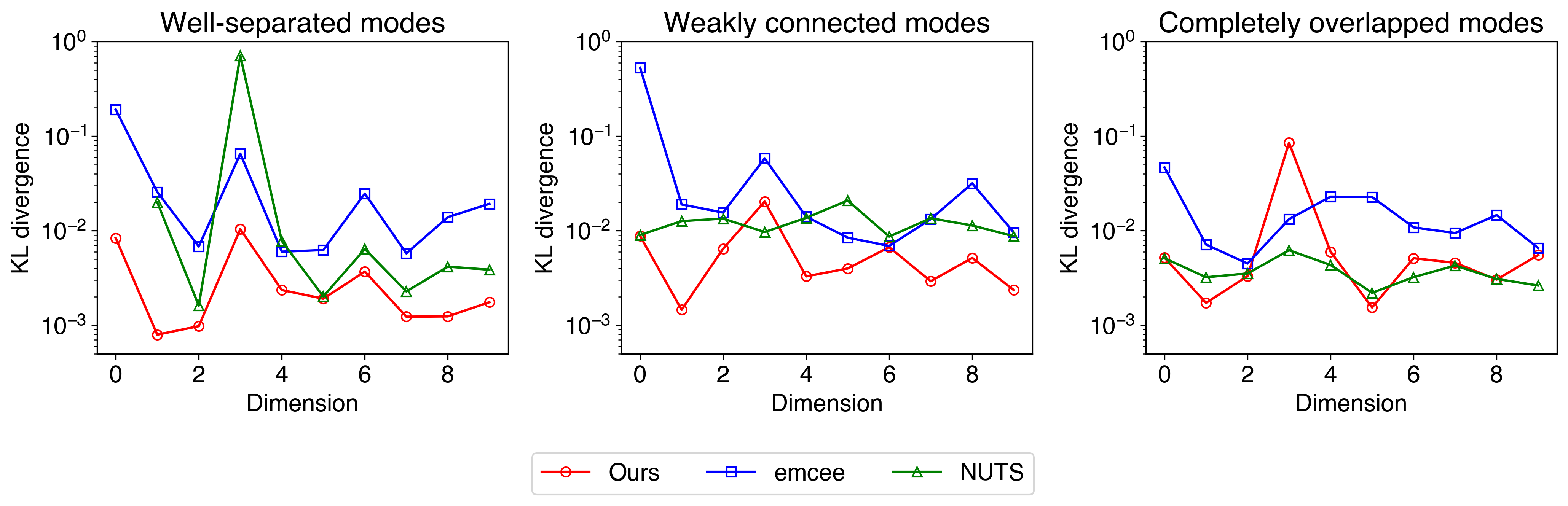}
\vspace{-.1in}
\caption{KL divergences between the $1d$ marginal distributions of samples from our framework and those from other baselines, relative to the ground truth marginal distributions in the first 10 dimensions. Our method achieves the lowest errors in both the well-separated and weakly connected case. In the completely overlapped setting, it performs on par with NUTS and outperforms Emcee, only slightly worse in the fourth dimension.}
\label{fig:test2_kl}
\end{figure}

Figure \ref{fig:test2_kl} displays the KL divergences for the $1d$ marginal distributions of our samples and those from other baselines over the first $10$ dimensions. In both the well-separated and weakly connected cases, our method achieves smallest errors, followed by NUTS and Emcee. All methods exhibit larger errors in the first dimension, where the target distribution is bimodal, and in the fourth dimension. The particularly high errors for NUTS and Emcee in the first dimension can be attributed to their inefficient exploration of the bimodal landscape (see Figure \ref{fig:test2_pdf100D}). In contrast, our framework consistently maintains errors below $10^{-2}$ in all dimensions. In the completely overlapped setting, our method mostly performs comparable to NUTS and is superior to Emcee, only slightly worse in the fourth dimension. Here, the KL divergences of all methods do not exceed $10^{-1}$, reflecting the unimodal and consequently easier-to-sample nature of the target distribution.

Table \ref{tab:test2:sinkhorn} reports the Sinkhorn distances computed between samples generated by our method, Emcee, and NUTS relative to the ``ground truth'' samples, with the Sinkhorn regularization parameter set to $0.05$. For the well-separated case ($\Delta =12$), our framework achieves the lowest distance, followed by Emcee and then NUTS. Emcee outperforms NUTS because it is able to capture the left mode -- albeit with some bias -- while NUTS entirely misses it (see Figure \ref{fig:test2_pdf100D}). For the weakly connected case ($\Delta =4$), our method and NUTS sample more accurately across the modes, resulting in smaller distances compared to Emcee, with our method taking the lead. For the completely overlapped case ($\Delta =0$), NUTS performs best, followed by our method and then Emcee, aligning with the KL divergence results for $1d$ marginal distributions (see Figure \ref{fig:test2_kl}).

To demonstrate the computational efficiency of our approach, Table \ref{tab:test2:running_time} compares the time required to generate new samples for our and other methods. Here, we focus on the sampling phase, i.e., after the generators are trained, thus excluding the training overhead for our framework. For a fair comparison with traditional probabilistic methods, we also omit their burn-in time. Additionally, since Emcee produces ten times as many samples as the other approaches (200000 vs. 20000), we divide its runtime time by ten to get an estimate of time cost per 20000 samples. All the sampling was performed on a 2.6 GHz 6-Core Intel Core i7 MacBook. Table \ref{tab:test2:running_time} shows that our method only requires roughly 2 seconds, half the time needed by Emcee, while NUTS as a gradient-based sampling approach is significantly more expensive. Also, the sampling time of our approach and Emcee does not change across three scenarios, while NUTS take double the time in the well-separated case.

\vspace{.2in}
\begin{minipage}{0.45\textwidth}
    \centering
    \begin{tabular}{cccc}
        \hline
        Cases & Ours & Emcee & NUTS \\ 
        \hline
        $\Delta=12$ & \textbf{11.21} & 12.66 & 13.21  \\  
        $\Delta=4$ & \textbf{11.65} & 11.92 & 11.75   \\  
        $\Delta=0$ & {11.50} & 11.64 & \textbf{11.42} \\  
        \hline
    \end{tabular}
    \captionof{table}{Sinkhorn distances (with regularization parameter $0.5$) for samples generated by our method,
Emcee, and NUTS relative to the ground truth. \textbf{Bold} indicates the best performance.}
    \label{tab:test2:sinkhorn}
\end{minipage}
\hfill
\begin{minipage}{0.45\textwidth}
    \centering
    \begin{tabular}{cccc}
        \hline
        Cases & Ours & Emcee &  NUTS \\ 
        \hline
        $\Delta=12$ & \textbf{2.18} & 4.23  & 47.61  \\  
        $\Delta=4$ & \textbf{2.21} & 4.18  & 22.30  \\  
        $\Delta=0$ & \textbf{2.17} & 4.26 & 20.84  \\  
        \hline
    \end{tabular}
    \captionof{table}{Comparison of runtime (in seconds) in sampling phase of our method and the other baselines. The generator training time (for our method) and burn-in time (for Emcee and NUTS) are excluded. }
        \label{tab:test2:running_time}
\end{minipage}



\subsection{$20$d skew-normal mixture model}
We consider a $20d$ mixture model composed of four skew-normal distributions, where each component represents a different mode of the overall distribution. The PDF of the mixture model is given by
\begin{equation}
\label{eq:sn_model1}
\rho({x}) = \sum_{k=1}^{4} r_k \mathcal{SN}({x} \mid {\mu}_k, {\Sigma}_k, {\alpha}_k),
\end{equation}
where $\mathcal{SN}({x} \mid {\mu}_k, {\Sigma}_k, {\alpha}_k)$ is the {$20d$ skew-normal density function with ${\mu}_k \in \mathbb{R}^{20}$ being the {mean vector} of the $k$-th component, ${\Sigma}_k \in \mathbb{R}^{20 \times 20}$ being the covariance matrix and ${\alpha}_k \in \mathbb{R}^{20}$ being the {skewness parameter} controlling the asymmetry of the distribution. In particular, each {$20d$ skew-normal distribution} follows the density function
\begin{equation}
\label{eq:sn_model2}
{\mathcal{SN}}({x} \mid {\mu}_k, {\Sigma}_k, {\alpha}_k) = 2 \mathcal{N}({x} \mid {\mu}_k, {\Sigma}_k) \Phi({\alpha}_k^\top  ({x} - {\mu}_k)),
\end{equation}
where $\mathcal{N}(\cdot \mid {\mu}, {\Sigma})$ is the {$20$-variate normal distribution} with mean ${\mu}$ and covariance ${\Sigma}$ and $\Phi(\cdot)$ is the cumulative distribution function of the univariate spherical Gaussian. The parameters of each mode are specified as
\begin{gather}
    \label{eq:20d_SN}
\begin{aligned}
\mu_1 &= (4,4,4,0,\ldots,0) ,\ \alpha_1 = (5,0,0,\ldots,0),\ \Sigma_1 = \begin{bmatrix} \hat{\Sigma}_1 & 0 \\ 0 & I_{18} \end{bmatrix},\ \text{with }\hat{\Sigma}_1 =  \begin{bmatrix} 1.5 & -0.9  \\ -0.9 & 1.5   \end{bmatrix}, \\
\mu_2 &= (-4,-4,4,0,\ldots,0) ,\ \alpha_2 = (-2,1,0,\ldots,0),\  \Sigma_2 = I_{20},\\
\mu_3 &= (-4,4,-4,0,\ldots,0) ,\ \alpha_3 = (5,0,0,\ldots,0),\ \Sigma_3 = \begin{bmatrix} \hat{\Sigma}_3 & 0 \\ 0 & I_{18} \end{bmatrix},\ \text{with }\hat{\Sigma}_3 =  \begin{bmatrix} 1.0 & 0.9  \\ 0.9 & 1.0   \end{bmatrix},\\
\mu_4 &= (4,-4,-4,0,\ldots,0) ,\ \alpha_4 = (5,5,0,\ldots,0),\  \Sigma_4 = I_{20}.
\end{aligned}
\end{gather}
The mixing coefficients are set to $r_1 = 0.35,\, r_2=0.27,\, r_3 = 0.17,\, r_4=0.21$. The distribution has four separated modes, with its projections to the first three dimensions displaying bimodality. 

The experimental setup of our methodology is detailed in Table \ref{tab:test3_hyperparameters}. We use a similar setup to the previous $100d$ Gaussian mixture test (see Section \ref{sec:test2_gaussian}), with slight adjustment on the prior domain, step size and numbers of SGD and Langevin iterations. For each component, we generate $5000$ samples via the Langevin dynamics and $5000$ labeled samples from the reverse ODE solutions, resulting a total of $20000$ samples. The final trained generative model is used to generate $20000$ samples of the target distribution for evaluation. For Emcee, we employ $60$ walkers -- three times the model dimension. Each walker is initialized by sampling from normal distribution with zero mean and standard deviation of $5$. We allow a burn-in period of $2000$ steps, followed by $500$ sampling steps per walker, resulting in a total of $60 \times 500 = 30000$ samples. For NUTS, we adopt a similar configuration as in our previous test: $20000$ burn-in steps and $20000$ sampling steps, with initial guesses drawn from a standard normal distribution, and a target mean acceptance probability set to $0.2$. In addition, we also include the dynamic nested sampling method \cite{higson2019dynamic}, which extends traditional nested sampling by dynamically allocating samples for improved efficiency, implemented with the Dynesty package \cite{sergey_koposov_2024_12537467, 10.1093/mnras/staa278}. We set the prior domain to $[-8,8]^{20}$, and adopt multi-ellipsoidal (\texttt{`multi'}) bounding method, as recommended for multimodal distributions. Since the analytic gradient of the target model is available, Hamitonian slice (\texttt{`hslice'})
as the sampling method.

Figure \ref{fig:test3_plot} displays a pair plot comparing the distributions of our samples, the ground truth, and those generated by other methods for the first five dimensions. The diagonal plots show the $1d$ marginal distribution of our samples together with the compared approaches. Note that the target model exhibits bimodality in the first three dimesions. Our method most accurately approximates the ground truth in all cases and significantly outperforms the other approaches. Dynesty is the second-best method; however, it fails to capture the correct mixing ratio and overestimates the support. In contrast, Emcee exhibits poor mixing—particularly in the second and third dimensions—while NUTS completely misses one mode in the first three dimensions, likely because it becomes trapped in a single mode of the target model. The off-diagonal plots provide the $2d$ joint distributions in each pair of dimensions among first five dimensions. Here, the lower-left subplots illustrate the joint distributions of our generated samples, while the upper-right sublots show corresponding distributions for the ground truth samples. We observe our approach fully captures all four modes of the target model with the correct sample allocation ratios. Moreover, it effectively reproduces the complex, skewed shapes of each local component, albeit with slightly less sharpness compared to the ground truth.

\begin{table}
   \centering
    \resizebox{0.65\textwidth}{!}{  
    \renewcommand{\arraystretch}{1.2} 
    \begin{tabular}{lc}
        \toprule
        \textbf{Step} & \textbf{Setup} \\
        \midrule
        \multirow{2}{*}{Step 1}  & 2000 starting points within \( [-8,8]^{20} \) \\
                                 & 10000 SGD iterations with step size \( \lambda = 0.02 \) \\
        \midrule
        \multirow{2}{*}{Step 2}  & 5000 samples per component \\
                                 & 40000 Langevin iterations with step size \( \eta = 0.002 \) \\
        \midrule
         \multirow{2}{*}{Step 3} &   explicit Euler scheme for reverse ODE with 100 time steps \\
                                & 5000 labeled samples generated per component \\
        \midrule
        \multirow{4}{*}{Step 4}  & Feedforward model with three hidden layers \\
                                 & Each layer has 1000 nodes with \texttt{tanh} activation \\
                                 & Adam with initial learning rate 0.001, halved every 500 epochs \\
                                 & Training over $2000$ epochs with batch size $1500$ \\
        \bottomrule
    \end{tabular}
    }
    \captionof{table}{Experimental setup of our framework for training a generative model on the $20d$ skew-normal mixture distribution \eqref{eq:sn_model1}-\eqref{eq:20d_SN}.}
    \label{tab:test3_hyperparameters}
\end{table}

\begin{figure}[!h]
\centering
\includegraphics[width = 0.9\textwidth]{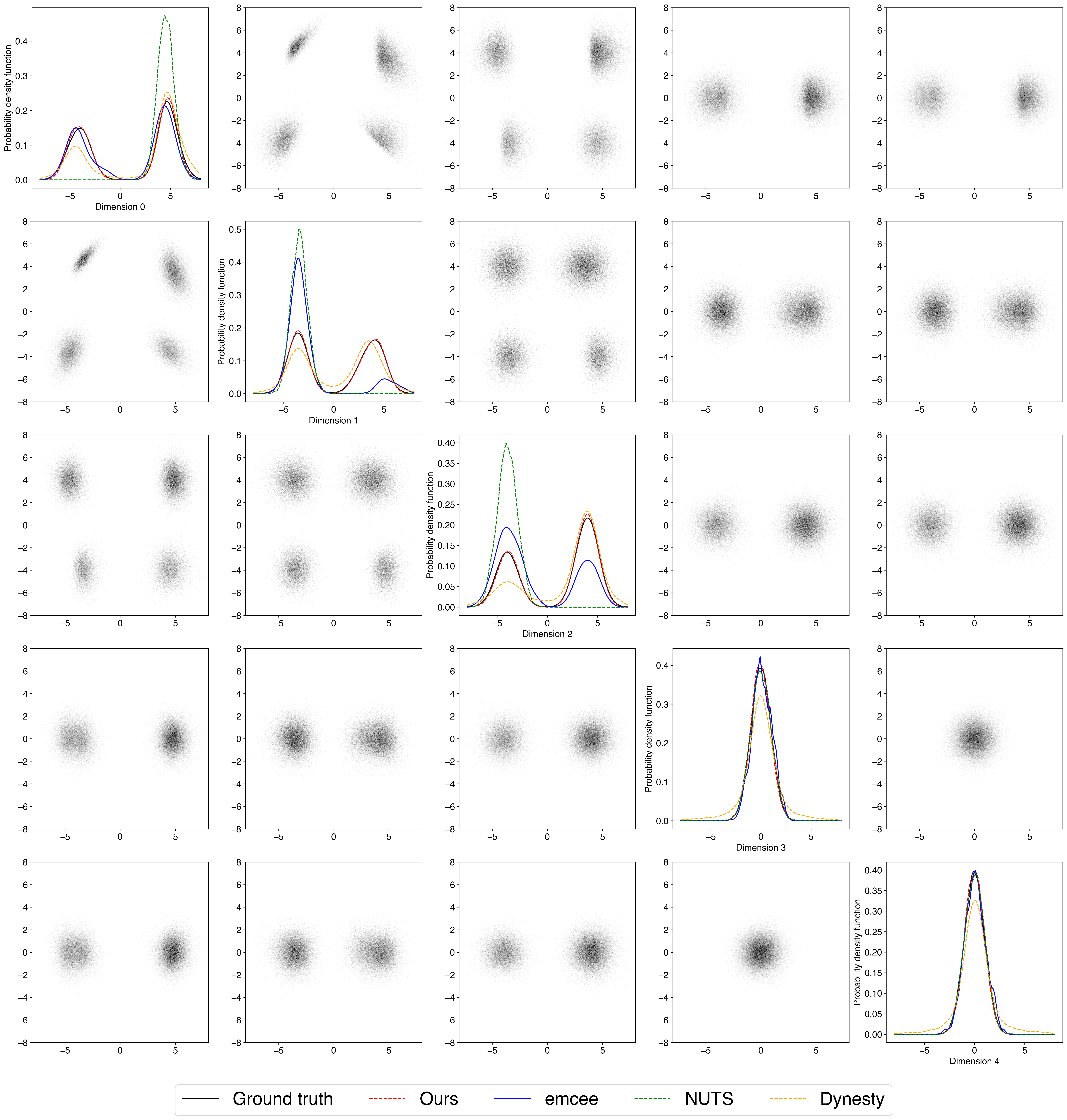}
\vspace{-.1in}
\caption{Comparison of samples from our approach, baselines, and the ground truth. Diagonal: $1d$ marginal distribution of samples in the first dimension. Our samples closely match the ground truth, whereas NUTS misses three of four modes, Emcee shows poor mixing, and Dynesty inaccurately estimates the mixing ratios and supports. Off-diagonal: $2d$ joint distributions of our generated samples (lower left) compared to the ground truth (upper right) in pair of dimensions among first five dimension. Our approach fully captures all four modes with correct sample allocation, though slightly less sharply than the ground truth.   }
\label{fig:test3_plot}
\end{figure}

Figure \ref{fig:test3_kl} presents the KL divergences for the $1d$ marginal distributions of our samples and those from the other approaches over all $20$ dimensions. In the first three dimensions, our method achieves smallest errors, consistent with Figure \ref{fig:test3_plot}, which shows it as the only method accurately capturing the bimodal marginal distributions. Dynesty and Emcee rank second and third, respectively, with Emcee performing particularly poorly in the second dimension. Although Dynesty outperforms Emcee and NUTS in the first three dimensions, its samples tend to exhibit overly wide support, resulting in less accurate marginal distributions in the remaining dimensions. On the other hand, NUTS fails to capture three out of the four modes of the target model, rendering its samples not useful, even though its marginal distributions appear the best from the fourth dimension onward.

\begin{figure}[!h]
\centering
\includegraphics[width = 0.8\textwidth]{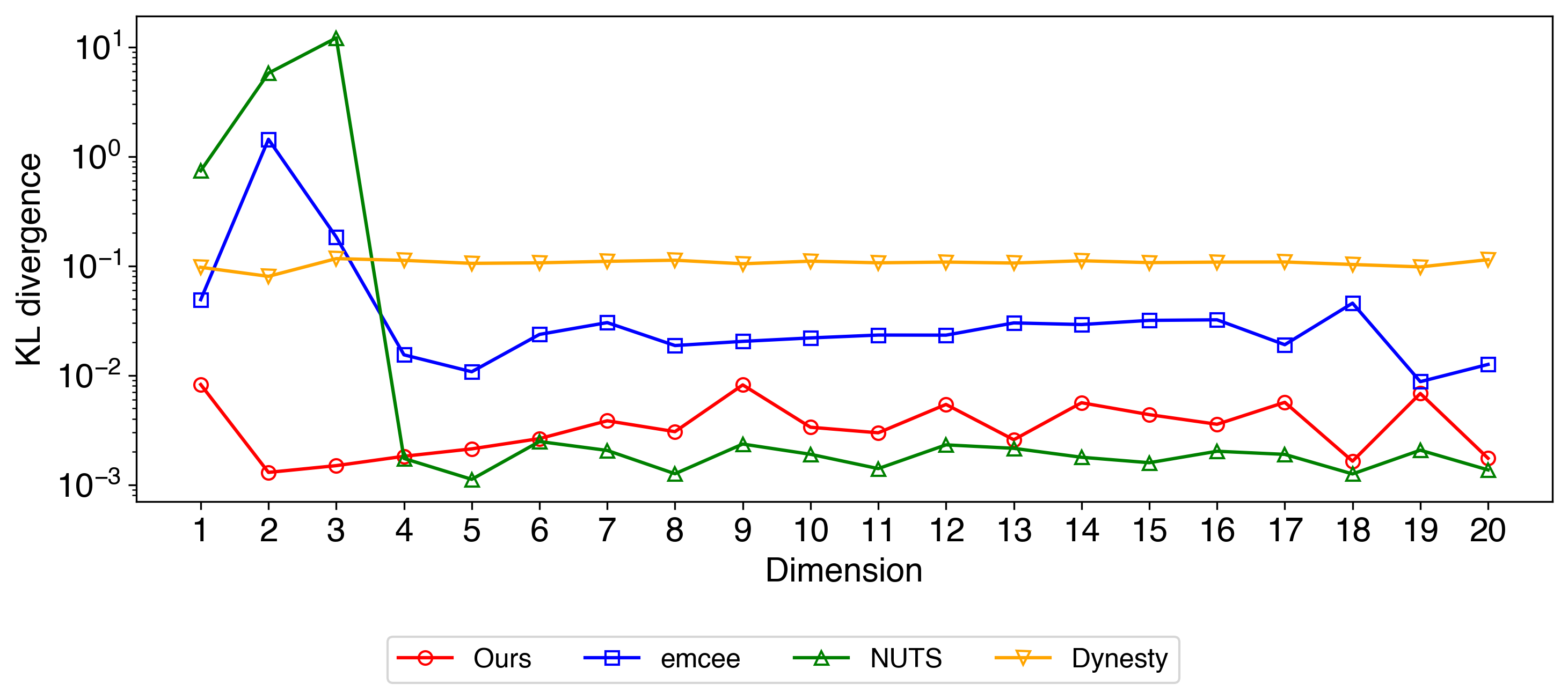}
\vspace{-.1in}
\caption{KL divergences for $1d$ marginals of samples from our framework and baselines relative to the ground truth marginals in all $20$ dimensions. Our method significantly outperforms the others in the first three dimensions, where the marginals exhibit bimodality. Although our method is second to NUTS from the fourth dimension onward, NUTS is trapped in a single mode, rendering its samples ineffective.}
\label{fig:test3_kl}
\end{figure}


Table \ref{tab:example} reports the Wasserstein distances between samples generated by our method, Emcee, NUTS, and Dynesty relative to the ground truth samples. Our framework achieves the smallest distance—four times lower than those obtained by Emcee and Dynesty—while NUTS performs the worst. We also compare the time required to generate new samples during the sampling phase. For Emcee, we further divide its runtime by $1.5$ to get an estimate of runtime per 20000 samples. Notably, our method requires less than one second, a fraction of the time needed by the other approaches. Perhaps because the mixture model is more complicated, the computation times for the baseline methods increase significantly compared to the previous test case. In contrast, our approach's sampling time relies solely on the trained generator and because we use a similar architecture in both cases, its runtime remains unchanged. Finally, while Dynesty ranks second in terms of Wasserstein distance, it requires considerably more time than the other baselines. 

\begin{table}[!h]
    \centering
    \begin{tabular}{ccccc}
        \hline
         & Ours & Emcee  & NUTS & Dynesty \\ 
        \hline
        Wasserstein distance & \textbf{13.70} & 58.69  & 103.60 & 55.45 \\  
        Runtime & \textbf{0.83s} & 113s  & $1074$s & $4$ hours \\  
        \hline
    \end{tabular}
    \caption{Comparison of Wasserstein distances and runtime in sampling phase of our approach versus baselines. Our framework shows a significant advantage in both metrics. }
    \label{tab:example}
\end{table}

\subsection{Highly complex densities from $2d$ images}
The effectiveness of our approach largely depends on the diffusion-based generative model for single-component sampling (Steps 3-4). In this test, we demonstrate that this module can function independently to generate samples from low-dimensional but highly complex distributions, showcasing the flexibility of our framework. We consider a problem from \cite{wu2020stochastic}, where three different images are used to define complex $2d$ functions as target densities to be sampled. Our approach is benchmarked against Emcee, Dynesty, and pure Metropolis Monte Carlo in generating 500000 samples from these target densities. Here, we estimate the score function using a Monte Carlo approximation with $20000$ samples drawn randomly from a uniform distribution over the image domain, rather than using Langevin dynamics as in the previous experiments. Then we obtain $500000$ labeled samples from the solution of reverse ODEs to train the generative model. The model architecture is similar to the previous tests, comprising a fully connected network with three hidden layers, each containing 500 nodes, and tanh activation functions. The final trained generative model is then employed to generate $500000$ new samples for evaluation. For Emcee, we employ $250$ walkers, each having a burn-in phase of $1000$ steps, followed by $2000$ sampling steps. The Dynesty method is tested with the default configuration, with multi-ellipsoidal (\texttt{`multi'}) bounding and random walk (\texttt{`rwalk'}) sampling strategy. In the Metropolis algorithm, we use 500000 independent chains, each running for 5000 steps with a step size of 0.01, and collect the final states of the chains as the sampled outputs.

The $2d$ histograms of generated samples are presented in Figure \ref{fig:test1b_compare}, along with their corresponding KL divergences from the ground truth densities. Here, the basic Metropolis struggles to sample accurately, with too many samples alloted to the background (low density regions) and insufficient samples at the important features. Dynesty also suffers from background oversampling, although to a lesser extent. The two top performers in this test are Emcee and our proposed method. Notably, our approach outperforms Emcee in the Dog and Smiley cases, ranking first in terms of KL divergence. However, it tends to over-smooth sharp edges, which makes it less effective in the Text case, where Emcee yields superior result.

\begin{figure}[!h]
\centering
\includegraphics[width = 0.9\textwidth]{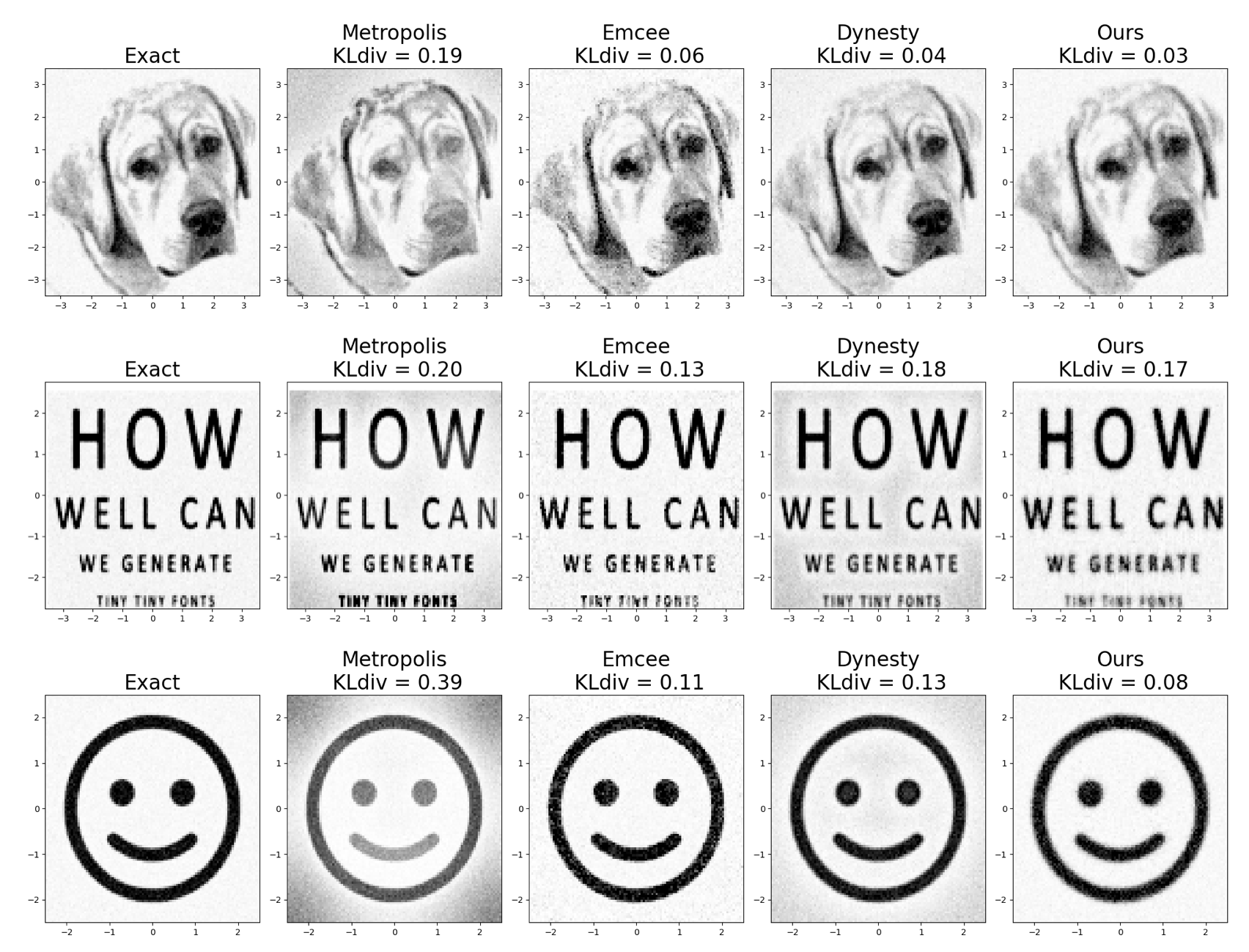}
\vspace{-.1in}
\caption{Sampling of Dog, Text and Smiley densities using our single component generative module (Step 3-4) compared to other approaches. `Exact' represents unbiased samples from exact density. Our approach achieves the best performance in Dog and Smiley cases, while for the Text density, where it ranks second to Emcee.}
\label{fig:test1b_compare}
\end{figure}

\subsection{Application to inverse partial differential equation (PDE)}

In this last test, we demonstrate an application of our framework to an inverse PDE problem. In particular, we consider the following parabolic equation modeling the contamination flows in the water resources research \cite{LI2015173, LIN2022111173}
\begin{gather}
\label{eq:PDE}
\begin{aligned}
    u_t &= \Delta u + f,\ x\in {\Omega},\, t\in [0,0.03], 
\\
u(x,0) & = \beta e^{-\|x-x_0\|^2/\alpha},
\\
u(x,t) & = v(x,t),\, x\in \partial \Omega. 
\end{aligned}
\end{gather}
Here, $\Omega = [0,1]\times[0,1]$, $u(x,t)$ is the concentration of the pollution at location $x$ and time $t$,  and $\alpha = 2h^2, \beta = M/(2\pi h^2)$ are two positive constants where $h$ is the radius of the pollution source and $M$ is the strength of the initial contamination. We set $h =0.1$ and $M=1$. The initial pollution source $x_0 \in \mathbb{R}^2$ is unknown and the objective of this problem is to trace back $x_0$
given some observed concentration $u(x,t)$ at some sensors in the domain. This is typically solved by the sampling methods in the Bayesian framework where the posterior PDF of $x_0$ is inferred by the discrepancy between the observed and simulated concentrations.
We set 
$
f = \beta e^{-\frac{\|x - x_0\|^2}{\alpha}} e^{-t} \left[-1 - \frac{4\|x - x_0\|^2}{\alpha^2} + \frac{4}{\alpha}\right]
$
so that the true solution of \eqref{eq:PDE} is $u(x,t) = \beta e^{-\|x-x_0\|^2/\alpha} e^{-t}$. As in \cite{LIN2022111173}, we consider two cases: 
\begin{enumerate}[label=\roman*.]
\item Two sensors are placed at $s_1 = (0.3,0.5)$ and $s_2 = (0.6,0.5)$ and sensor measurements of the concentration at the terminal time $t = 0.03$ are $u(s_1,t)= u(s_2,t)= e^{-r^2/\alpha} \cdot \beta \cdot e^{-0.03}$, where $r = 0.2$. This setting has two inverse solutions; 
\item One sensor is placed at $s_1 = (0.3,0.5)$. In this scenario, the inverse target forms a ring centered at the sensor, resulting in an infinite number of possible solutions to the inverse problem.
\end{enumerate}
The primary challenge in this problem is that each evaluation of the target PDF requires solving a forward PDE, thus limiting the number of allowable PDF calls. Here, we set the maximum number of PDF calls to $20000$ and assess the performance of our proposed method in comparison to Metropolis Monte Carlo and Dynesty. We omit Emcee which tends to diverge in this setting, as well as NUTS, since it requires gradient information that is not available. We set the prior domain for $x_0$ to be $[0,0.8 ]\times [0,0.8]$. For our approach, we only use the diffusion-based generative module for single-component sampling. The score function is estimated using Monte Carlo approximation with samples drawn randomly from a uniform distribution over the prior domain. This is the only step where PDF queries are used, and no additional PDF evaluations are needed thereafter in our framework. We use the same model architecture as in the $2d$ image experiment. The Dynesty method is tested with the default configuration, with multi-ellipsoidal (\texttt{`multi'}) bounding and random walk (\texttt{`rwalk'}) sampling strategy. For the Metropolis algorithm, we use a single chain running for 20000 steps with a step size of 0.02. 

In Figures \ref{fig:PDE1}-\ref{fig:PDE2}, we present the scatter plots of the proposed samples for both cases. The proposed Metropolis samples are collected after convergent. For Dynesty, we only plot the effective samples, which constitute about $3$-$10\%$ of the PDF calls. For our method, we employ the final trained generative model to create $2500$ new samples for evaluation. The plots illustrate that Metropolis struggles to fully capture the PDF modes, as it misses one mode in the first case and captures only part of the PDF support in the second case. Meanwhile, Dynesty successfully identifies the modes but fail to emphasize their significance and places an excessive number of samples in the low density region. In comparison, our approach accurately captures the target PDF, effectively highlighting the modes without overrepresenting low-density areas. These results demonstrate the clear advantage of our method over the baseline approaches.

\begin{figure}[!h]
\centering
\includegraphics[width = \textwidth]{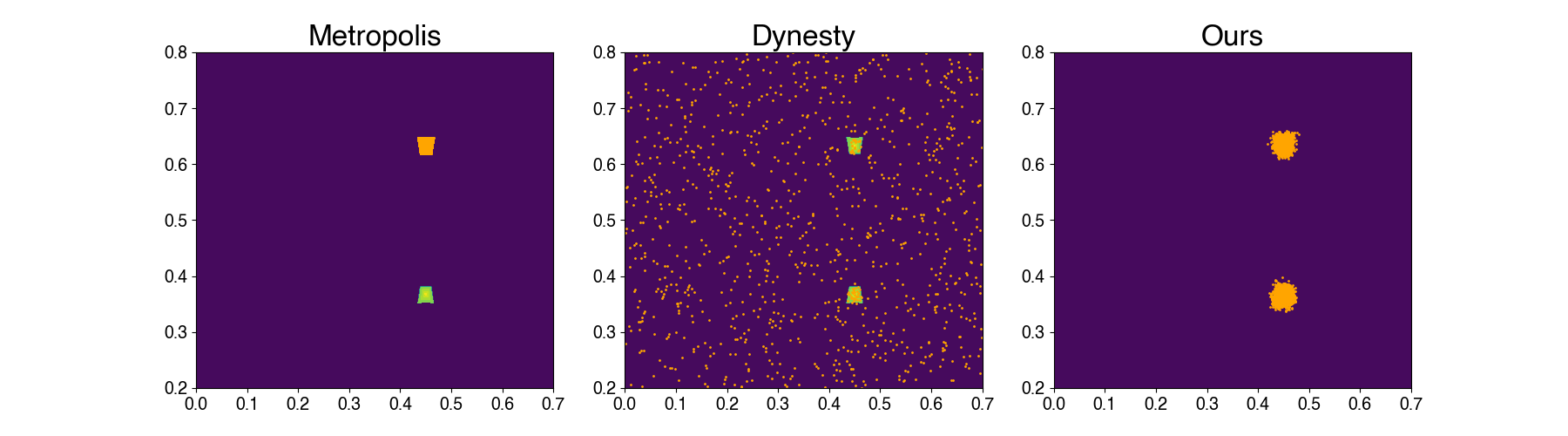}
\vspace{-.3in}
\caption{Scatter plot of the proposed samples for Case i. The
light green regions indicate the potential exact locations of the pollution sources, and the orange dots are the proposed samples. Metropolis only captures one of the modes (left), whereas Dynesty successfully identifies both modes but fails to emphasize their significance and places a large number of samples in the low density region (center). In contrast, our approach can capture both modes correctly (right).    }
\label{fig:PDE1}
\end{figure}

\begin{figure}[!h]
\centering
\includegraphics[width = \textwidth]{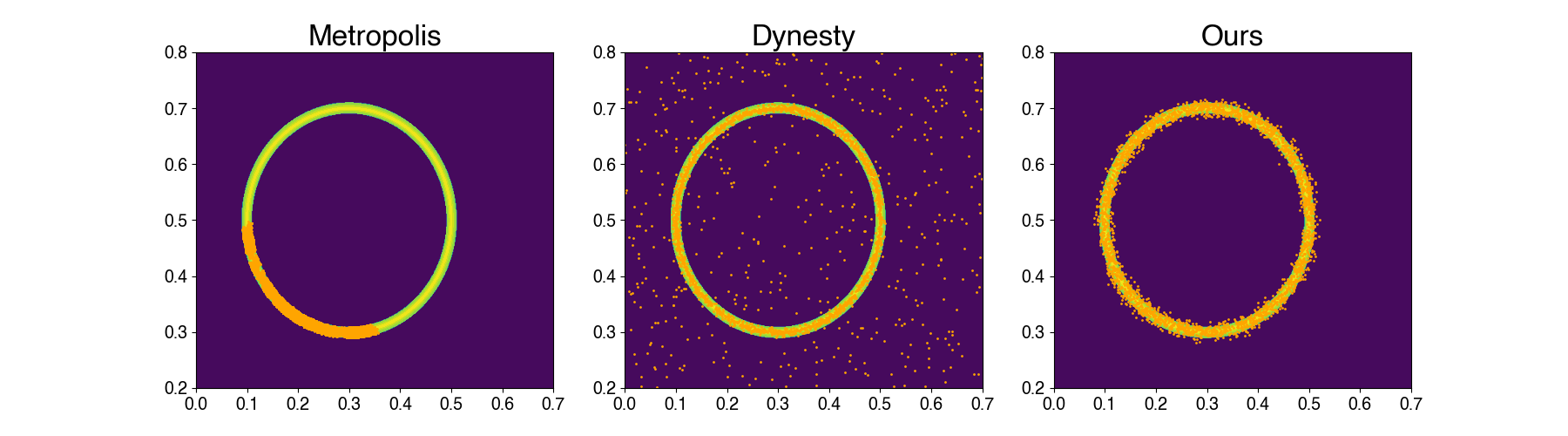}
\vspace{-.3in}
\caption{Scatter plot of the proposed samples for Case ii. The target PDF has a ring shape. Metropolis captures only part of the ring due to an insufficient number of samples (left), while Dynesty successfully identifies the ring but oversamples the low density region (right). In contrast, our approach can capture the ring accurately (right). }
\label{fig:PDE2}
\end{figure}

\section{Conclusions}\label{sec:conclusion}

We develop a novel framework for supervised learning of generative models for high-dimensional, multimodal distributions. The key idea of our approach is divide-and-conquer strategy, which breaks the sampling task into smaller, more manageable subproblems. For each subproblem, we develop a training-free diffusion model to label and then generate samples of the target PDF from the standard Gaussian samples. Our method shares the advantage of state-of-the-art deep-learning-based approaches, such as normalizing flows and diffusion models, enabling near-instantaneous generation of samples after an offline training phase. However, it does not require complex architecture designs, as we demonstrate that a simple fully connected network with three hidden layers is sufficient to handle all test cases presented in this paper. The framework is modular and flexible, comprising several components that can be adapted and even function independently based on specific application needs. Furthermore, improvements to any individual component directly enhance the overall performance of our framework, paving the way for numerous future research directions.

\section*{Acknowledgement}
This work is supported by the U.S. Department of Energy (DOE), Office of Science, Office of Advanced Scientific Computing Research, Applied Mathematics program, under the contract ERKJ443. It is also supported by Dan Lu’s Early Career Project, sponsored by the Office of Biological and Environmental Research in DOE. All the work was accomplished at Oak Ridge National Laboratory (ORNL) which is operated by UT-Battelle, LLC., for the DOE under Contract DE-AC05-00OR22725. The third author (FB) would also like to acknowledge the support from U.S. National Science Foundation through project DMS-2142672 and the support from the U.S. Department of Energy, Office of Science, Office of Advanced Scientific Computing Research, Applied Mathematics program under Grant DE-SC0022297.

\bibliographystyle{siam}
\bibliography{Reference,nfref,GZ_ref}

\begin{thebibliography}{10}

\bibitem{BENNETT1976245}
{\sc C.~H. Bennett}, {\em Efficient estimation of free energy differences from
  monte carlo data}, Journal of Computational Physics, 22 (1976), pp.~245--268.

\bibitem{Buchner2021}
{\sc J.~Buchner}, {\em Ultranest - a robust, general purpose bayesian inference
  engine}, Journal of Open Source Software, 6 (2021), p.~3001.

\bibitem{8253599}
{\sc A.~Creswell, T.~White, V.~Dumoulin, K.~Arulkumaran, B.~Sengupta, and A.~A.
  Bharath}, {\em Generative adversarial networks: An overview}, IEEE Signal
  Processing Magazine, 35 (2018), pp.~53--65.

\bibitem{du2019implicit}
{\sc Y.~Du and I.~Mordatch}, {\em Implicit generation and modeling with energy
  based models}, Advances in Neural Information Processing Systems, 32 (2019).

\bibitem{earl2005parallel}
{\sc D.~J. Earl and M.~W. Deem}, {\em Parallel tempering: Theory, applications,
  and new perspectives}, Physical Chemistry Chemical Physics, 7 (2005),
  pp.~3910--3916.

\bibitem{feroz2009multinest}
{\sc F.~Feroz, M.~Hobson, and M.~Bridges}, {\em Multinest: an efficient and
  robust bayesian inference tool for cosmology and particle physics}, Monthly
  Notices of the Royal Astronomical Society, 398 (2009), pp.~1601--1614.

\bibitem{Emcee}
{\sc D.~{Foreman-Mackey}, D.~W. {Hogg}, D.~{Lang}, and J.~{Goodman}}, {\em
  emcee: The mcmc hammer}, PASP, 125 (2013), pp.~306--312.

\bibitem{4767596}
{\sc S.~Geman and D.~Geman}, {\em Stochastic relaxation, gibbs distributions,
  and the bayesian restoration of images}, IEEE Transactions on Pattern
  Analysis and Machine Intelligence, PAMI-6 (1984), pp.~721--741.

\bibitem{goodman2010ensemble}
{\sc J.~Goodman and J.~Weare}, {\em Ensemble samplers with affine invariance},
  Communications in applied mathematics and computational science, 5 (2010),
  pp.~65--80.

\bibitem{higson2019dynamic}
{\sc E.~Higson, W.~Handley, M.~Hobson, and A.~Lasenby}, {\em Dynamic nested
  sampling: an improved algorithm for parameter estimation and evidence
  calculation}, Statistics and Computing, 29 (2019), pp.~891--913.

\bibitem{NEURIPS2020_4c5bcfec}
{\sc J.~Ho, A.~Jain, and P.~Abbeel}, {\em Denoising diffusion probabilistic
  models}, in Advances in Neural Information Processing Systems, vol.~33,
  Curran Associates, Inc., 2020, pp.~6840--6851.

\bibitem{10.5555/2627435.2638586}
{\sc M.~D. Homan and A.~Gelman}, {\em The no-u-turn sampler: adaptively setting
  path lengths in hamiltonian monte carlo}, J. Mach. Learn. Res., 15 (2014),
  p.~1593–1623.

\bibitem{kobyzev2020normalizing}
{\sc I.~Kobyzev, S.~J. Prince, and M.~A. Brubaker}, {\em Normalizing flows: An
  introduction and review of current methods}, IEEE transactions on pattern
  analysis and machine intelligence, 43 (2020), pp.~3964--3979.

\bibitem{sergey_koposov_2024_12537467}
{\sc S.~Koposov, J.~Speagle, K.~Barbary, G.~Ashton, E.~Bennett, J.~Buchner,
  C.~Scheffler, B.~Cook, C.~Talbot, J.~Guillochon, P.~Cubillos, A.~A. Ramos,
  M.~Dartiailh, Ilya, E.~Tollerud, D.~Lang, B.~Johnson, jtmendel, E.~Higson,
  T.~Vandal, T.~Daylan, R.~Angus, patelR, P.~Cargile, P.~Sheehan, M.~Pitkin,
  M.~Kirk, J.~Leja, joezuntz, and D.~Goldstein}, {\em joshspeagle/dynesty:
  v2.1.4}, June 2024.

\bibitem{lecun2006tutorial}
{\sc Y.~LeCun, S.~Chopra, R.~Hadsell, M.~Ranzato, F.~Huang, et~al.}, {\em A
  tutorial on energy-based learning}, Predicting structured data, 1 (2006).

\bibitem{LI2015173}
{\sc W.~Li and G.~Lin}, {\em An adaptive importance sampling algorithm for
  bayesian inversion with multimodal distributions}, Journal of Computational
  Physics, 294 (2015), pp.~173--190.

\bibitem{LIN2022111173}
{\sc G.~Lin, Y.~Wang, and Z.~Zhang}, {\em Multi-variance replica exchange
  sgmcmc for inverse and forward problems via bayesian pinn}, Journal of
  Computational Physics, 460 (2022), p.~111173.

\bibitem{JMLMC_2024_Diffusion}
{\sc Y.~Liu, M.~Yang, Z.~Zhang, F.~Bao, Y.~Cao, and G.~Zhang}, {\em
  Diffusion-model-assisted supervised learning of generative models for density
  estimation}, Journal of Machine Learning for Modeling and Computing, 5
  (2024), pp.~25--38.

\bibitem{lu2022dpmsolver}
{\sc C.~Lu, Y.~Zhou, F.~Bao, J.~Chen, C.~Li, and J.~Zhu}, {\em {DPM}-solver: A
  fast {ODE} solver for diffusion probabilistic model sampling in around 10
  steps}, in Advances in Neural Information Processing Systems, A.~H. Oh,
  A.~Agarwal, D.~Belgrave, and K.~Cho, eds., 2022.

\bibitem{meng1996simulating}
{\sc X.-L. Meng and W.~H. Wong}, {\em Simulating ratios of normalizing
  constants via a simple identity: a theoretical exploration}, Statistica
  Sinica,  (1996), pp.~831--860.

\bibitem{10.1063/1.1699114}
{\sc N.~Metropolis, A.~W. Rosenbluth, M.~N. Rosenbluth, A.~H. Teller, and
  E.~Teller}, {\em Equation of state calculations by fast computing machines},
  The Journal of Chemical Physics, 21 (1953), pp.~1087--1092.

\bibitem{neal2011mcmc}
{\sc R.~M. Neal}, {\em Mcmc using hamiltonian dynamics}, in Handbook of Markov
  Chain Monte Carlo, Chapman and Hall/CRC, 2011, pp.~113--162.

\bibitem{platt1999probabilistic}
{\sc J.~PLATT}, {\em Probabilistic outputs for support vector machines and
  comparisons to regularized likelihood methods}, Advances in Large Margin
  Classifiers,  (1999), pp.~61--74.

\bibitem{pmlr-v115-song20a}
{\sc Y.~Song, S.~Garg, J.~Shi, and S.~Ermon}, {\em Sliced score matching: A
  scalable approach to density and score estimation}, in Proceedings of The
  35th Uncertainty in Artificial Intelligence Conference, R.~P. Adams and
  V.~Gogate, eds., vol.~115 of Proceedings of Machine Learning Research, PMLR,
  22--25 Jul 2020, pp.~574--584.

\bibitem{song2021scorebased}
{\sc Y.~Song, J.~Sohl-Dickstein, D.~P. Kingma, A.~Kumar, S.~Ermon, and
  B.~Poole}, {\em Score-based generative modeling through stochastic
  differential equations}, in International Conference on Learning
  Representations, 2021.

\bibitem{10.1093/mnras/staa278}
{\sc J.~S. Speagle}, {\em dynesty: a dynamic nested sampling package for
  estimating bayesian posteriors and evidences}, Monthly Notices of the Royal
  Astronomical Society, 493 (2020), pp.~3132--3158.

\bibitem{swendsen1986replica}
{\sc R.~H. Swendsen and J.-S. Wang}, {\em Replica monte carlo simulation of
  spin-glasses}, Physical review letters, 57 (1986), p.~2607.

\bibitem{10.1162/NECO_a_00142}
{\sc P.~Vincent}, {\em A connection between score matching and denoising
  autoencoders}, Neural Comput., 23 (2011), p.~1661–1674.

\bibitem{Wang2016WarpBS}
{\sc L.~Wang, D.~E. Jones, and X.-L. Meng}, {\em Warp bridge sampling: The next
  generation}, Journal of the American Statistical Association, 117 (2016),
  pp.~835 -- 851.

\bibitem{welling2011bayesian}
{\sc M.~Welling and Y.~W. Teh}, {\em Bayesian learning via stochastic gradient
  langevin dynamics}, in Proceedings of the 28th international conference on
  machine learning (ICML-11), Citeseer, 2011, pp.~681--688.

\bibitem{wu2020stochastic}
{\sc H.~Wu, J.~K{\"o}hler, and F.~No{\'e}}, {\em Stochastic normalizing flows},
  Advances in Neural Information Processing Systems, 33 (2020), pp.~5933--5944.

\bibitem{10.1145/3626235}
{\sc L.~Yang, Z.~Zhang, Y.~Song, S.~Hong, R.~Xu, Y.~Zhao, W.~Zhang, B.~Cui, and
  M.-H. Yang}, {\em Diffusion models: A comprehensive survey of methods and
  applications}, ACM Comput. Surv.,  (2023).
\newblock Just Accepted.

\end{thebibliography}

\appendix

\end{document}